\documentclass{amsart}
\usepackage{algorithmic}
\usepackage{algorithm}
\usepackage{graphicx}
\usepackage{subfigure}
\usepackage[toc,page]{appendix}
\usepackage{multirow}

\newtheorem{proposition}{Proposition}
\newtheorem{theorem}{Theorem}

\title{Efficient Multi-Template Learning for Structured Prediction}
\author[Q. Mao]{Qi Mao}
\author[Ivor W. Tsang]{Ivor W. Tsang}
\thanks{Qi Mao and Ivor W. Tsang are with School of Computer Engineering,
Nanyang Technological University, Singapore 639798, e-mail \{QMAO1,IvorTsang\}@ntu.edu.sg}

\begin{document}

\maketitle

\begin{abstract}
Conditional random field (CRF) and Structural Support Vector Machine (Structural SVM) are two state-of-the-art methods for structured prediction which captures the interdependencies
among output variables. The success of these methods is attributed to the fact that their discriminative models are able to account for overlapping features on the whole input
observations. These features are usually generated by applying a given set of templates on labeled data, but improper templates may lead to degraded performance. To alleviate this
issue, in this paper, we propose a novel multiple template learning paradigm to learn structured prediction and the importance of each template simultaneously,
so that hundreds of arbitrary templates could be added into the learning model without caution. This paradigm can be formulated as a special multiple kernel learning problem with exponential number of constraints. Then we introduce an efficient cutting plane algorithm to solve this problem in the primal, and its convergence is  presented.
We also evaluate the proposed learning paradigm on two widely-studied structured prediction tasks, \emph{i.e.} sequence labeling and dependency parsing.
Extensive experimental results show that the proposed method outperforms CRFs and Structural SVMs due to exploiting the importance of each template.
Our complexity analysis and empirical results also show that our proposed method is more efficient than OnlineMKL on very sparse and high-dimensional data. We further extend this paradigm for structured prediction using generalized $p$-block norm regularization with $p>1$, and experiments show competitive performances when $p \in [1,2)$.

\end{abstract}

Structured prediction \cite{Lafferty01,Taskar03,Tsochantaridis05} has been successfully applied to the problems with strong interdependencies among the output variables. In the realm of Natural Language Processing (NLP), various tasks are formulated into structured prediction problems. A typical example is part-of-speech tagging which assigns a specific part-of-speech tag to each token of an input sentence. The tag of one token is strongly correlated with the tags of its neighbors under the linear chain dependencies~\cite{Lafferty01,Tsochantaridis05}. More complicated structured output dependencies could be trees or graphs, such as Context-Free Grammar (CFG)~\cite{Tsochantaridis05}, dependency parsing tree~\cite{McDonald05a,McDonald05}, noun phrase coreference \cite{Finley2005}, and factor graph for relation extraction~\cite{Yao10}. Note that there exist exact inference methods for sequences and trees. For the tasks with general output structures   (\emph{eg.},  the pairwise fully connected undirected graph) the exact inference problem is intractable. In such cases, approximate inference is usually pursued to obtain an approximate solution~\cite{Finley08}.

The major advantage of structured prediction models such as Conditional Random Fields (CRFs) \cite{Lafferty01} and Structural Support Vector Machines (Structural SVMs) \cite{Taskar03,Tsochantaridis05} is that their learning models can easily integrate prior knowledge of a specific domain by feature engineering. For example, the discriminative models of CRFs can account for overlapping features (\emph{eg.}, first-order or even higher-order linear chain) on the whole observation sequence \cite{Lafferty01}. On the other hand,
 Structural SVMs \cite{Tsochantaridis05,Joachims09} relies on joint feature maps over the input-output pairs, where features can be represented equivalently as that of CRFs.

During last decade, structured prediction algorithms take more effort on how to model the interdependencies among the output variables, but less consideration is taken on the  feature engineering  which is a non-trivial and tedious task for general users. We observe that different kinds of rules are used to extract features from input sentences in both  sequence learning \cite{Peng04,McCallum03} and  dependency parsing \cite{McDonald05a}. The arbitrary non-independent features are usually extracted from a given input-output pair using a set of predefined templates (rules or feature functions). Templates can be arbitrarily defined according to specific applications  by exploring any internal or external knowledge as much as possible, and then features extracted from different templates are concatenated into the learning models so as to boost prediction performance.  However, features generated from arbitrary templates may be redundant or non-informative.  Structured prediction models, such as CRFs and Structural SVMs, only treat the features generated from each template equally without exploiting the importance of each template or its generated features. Therefore, some improper templates may generate conflicting or noisy features which  degrade these structured prediction models.

In order to prune noisy or non-informative features for improved generalization and fast prediction, feature selection strategy can be incorporated into the structural prediction models in two directions. One way is to treat each instantiated feature from each template independently. For example, CRFs with $\ell_1$ regularization has been explored for activity recognition by  \cite{Vail07}. However, progressively reducing the features by $\ell_1$ regularization methods may degrade the performance vastly due to the interdependencies among features in the same group.
Another way is to consider that features are naturally grouped together by templates \cite{Martins10}, and each group of features forms a base kernel in the framework of Multiple Kernel Learning (MKL)~\cite{Lanckriet04}. As discussed in~\cite{Bach04,Bach08},  group lasso and MKL can be used to obtain the group sparsity via the $\ell_{2,1}$ mixed norm and its square, respectively. More general groups sparsity regularizer such as $\ell_p$-$\ell_q$ has also been explored \cite{Rakotomamonjy2011} where $0 \leq p \leq 1$ and $1 \leq q \leq 2$. There is a plethora of note worthy research progress made in  group lasso and MKL. Many works~\cite{Sonnenburg06,Rakotomamonjy08,Bach08,Xu09,Tomioka09,Hu2009,Yang2011} focus on convergence and scalability issues on their optimization.
Zien and Ong~\cite{Zien07} formulated MKL for multiple classification problem in terms of structural prediction, but only focused on multiple class classification. Although SVM-based wrapped methods~\cite{Sonnenburg06,Rakotomamonjy08,Xu09} with SMO solver~\cite{Lopez2012} can efficiently solve some specific formulations of MKLs, it is intractable for these MKL and group lasso methods to deal with structured prediction due to the exponential number of constraints. Instead of the natural groups formed over features, we also notice that the training data can also be naturally separated into groups, known as ``Learning from structured data''~\cite{Cai2012}. However, groups in the feature level is the focus of this paper.

Recently, Martins \emph{et al.}~\cite{Martins10} proposed an online multi-kernel learning strategy, namely OnlineMKL,  to cater for the constraints of structure prediction. Recall that, due to the arbitrary design of templates in NLP, the extracted features are usually very high-dimensional and very sparse.
 Though the gradient descent step in each round of OnlineMKL training is  efficient, the subsequent proximal projection step on the dense and very high dimensional weight vector of the learning model
 is  very time-consuming (see Section~\ref{sec:complexity}).
Apart from these computational issues, Dekel and Singer~\cite{Dekel05} stated that there is no notion of statistical generalization in the online learning setting, so additional online-to-batch strategy is required to convert online classifiers for the batch-mode learning setting which fits
 most real-world problems more naturally~\cite{Dekel2008,Dekel05}.
Even worse, it is nontrivial to determinate the termination condition for online methods in order to achieve the optimal batch-mode prediction performance.

To overcome above problems, in this paper, we propose a Multiple Template Learning (MTL$^{struct}$) paradigm to learn the weight of each template and the structured prediction model in the batch mode, simultaneously. As noticed that MTL is also the abbreviation of Multi-Task Learning ~\cite{Caruana1997}. In what follows, MTL only stands for Multiple Template learning.
Following the group sparsity, to capture the dependency among a group of features instantiated from one template,
 learning the weighting of these groups is formulated as a Multiple Kernel Learning (MKL) problem.
 This specific MKL problem usually involves the exponential number of constraints due to the interdependencies among output variables. We propose to solve this MKL problem in the primal by an efficient cutting plane algorithm.
 Moreover, its convergence is presented. The main contributions of this paper are listed as follows,
\begin{itemize}
\item Following the idea of group feature selection (template selection), we propose to learn the large margin based structured prediction model in the batch mode. Hence, the generalization performance can be guaranteed. As opposed to OnlineMKL, the empirical risk is naturally considered as the criterion of termination condition for our proposed MTL$^{struct}$.
\item An adapted cutting plane algorithm is proposed to tackle the MKL problem with exponential number of constraints, and the convergence is guaranteed. This also leads to very sparse solution of the learned weights of templates, which is effective to prune noisy or non-informative templates for performance improvement, faster prediction and enhancing semantic interpretation.
\item Detailed time complexities of  state-of-the-art structured prediction methods are presented.
 These reveal that the proposed MTL$^{struct}$, which avoids the proximal projection on the dense weight vector,
 is much more efficient than OnlineMKL in the case of very sparse and high-dimensional data, which is common in NLP applications when hundreds of templates are used. And the training efficiency of the proposed MTL$^{struct}$ is also very comparable with Structural SVM. Moreover, since MTL$^{struct}$ can automatically remove redundant templates, the testing efficiency of MTL$^{struct}$ is much improved over structured SVM and various CRFs without performance reduction.
\item Two well-known structured prediction tasks, i,e. sequence labeling and dependency parsing, are showcased in this paper.
    Extensive experimental results demonstrate that the proposed paradigm is more effective than OnlineMKL to learn the importance of each template for the structured prediction tasks.  The learned weights have the ability to avoid the degraded performance which is caused by adding poorly designed or even conflicting templates into the learning model. This phenomenon is demonstrated in the experiments, so  our proposed MTL$^{struct}$ can tackle the challenge of choosing and weighting the informative templates for the optimal prediction performances.
\item We further extend the proposed learning paradigm to solve structured output problems with $p$-block norm regularization. Experimental results on sequence labeling tasks show competitive results when $p\in[1,2)$; otherwise, the overall performance may degrade vastly.
\end{itemize}

The rest of this paper is organized as follows. The proposed MTL$^{struct}$ framework for structured prediction is presented in Section~\ref{sec:MTL} and its extension to $p$-block norm regularization is given in Section~\ref{sec:pmkl}. Section~\ref{sec:experiments} presents the experimental results of two showcases, i.e, sequence labeling and dependency parsing. Conclusive remarks are given in Section~\ref{sec:conclusion}.

\vspace{-0.1in}
\section{Multiple Template Learning} \label{sec:MTL}

CRFs and Structural SVMs have already been used to model many structured prediction problems, such as sequence labeling with a first-order or even higher-order linear chain, and syntactic parsing with a directed tree. More general structures are possible if a feasible inference method exists. CRFs are graphical models by defining conditional probability density function over the given graph structure; while Structural SVMs incorporate the structural information by defining joint feature mapping under large margin theory. Even though two models are derived under different principles, they all confront the same challenge: how to extract features from structured input-output pairs so as to obtain better prediction performance.  While our MTL$^{struct}$ framework mainly focuses on how to efficiently and effectively manage the given set of templates and learn a better structured prediction model, simultaneously.

\vspace{-0.1in}
\subsection{Feature v.s. Template} \label{sec:template}

For the problems with structured inputs, features are usually not explicitly defined. For instance, part-of-speech tagging in  NLP is to label each token with a specific tag in the given token sequence (sentence). In addition to the current token, the neighbors and their associated tags can be considered as the important features to determine the tag of the current token. All these intuitive information can be represented by feature functions called templates \cite{Peng04,McCallum03,McDonald05a}. Hereby, we consider the templates as some predefined rules for feature extraction.

Given a training dataset $\mathcal{D}=\{ (\mathcal{X}_i,\mathcal{Y}_i)\}_{i=1}^n$ with $n$ structured input-output pairs $(\mathcal{X}_i, \mathcal{Y}_i )$ where $\mathcal{X}_i \in \mathbb{X}$ and $\mathcal{Y}_i \in \mathbb{Y}$ can be any structural objects, for instance, sequences, trees, or general graphs. Assume that there are $m$ templates denoted by the operator $\kappa_j(.),\forall j=1,\ldots,m$ over the domain of structural input-output pairs to generate a list of features, e.g., the bigram rule in the part of speech tagging.  By applying $\kappa_j(.)$ to all pairs of $\mathcal{D}$, we can instantiate a set $\kappa_j(\mathcal{D})=\{\kappa_j^1,\ldots,\kappa_j^{d_j}\}$ with $d_j$ different features. Therefore, the set of features over $\mathcal{D}$ are $\kappa(\mathcal{D}) = \{ \kappa_1(\mathcal{D}),\ldots,\kappa_m(\mathcal{D})\}$. The size of total features $|\kappa(\mathcal{D})|$ may be smaller than $\sum_{j=1}^m |\kappa_j(\mathcal{D})|$ since one feature may be generated from more than one template.
The case of one feature in more than one group is not the concern of this paper,
so we augment the index of the applied templates to the instantiated feature such that one feature belongs to only one group.
Each input-output pair $(\mathcal{X}_i, \mathcal{Y}_i)$ now can be represented by a real value feature vector $\Phi(\mathcal{X}_i, \mathcal{Y}_i)$ using the same set of features in $\kappa(\mathcal{D})$. Since $\kappa(\mathcal{D})$ is extracted from $\mathcal{D}$ and a small subset of features in $\Phi(\mathcal{X}_i, \mathcal{Y}_i)$ could be activated, so each instance may have a very sparse feature representation.

Previous methods for structured inputs and outputs did not consider the properties of features, and directly used $\kappa(\mathcal{D})$ as the feature representation. Taking part-of-speech tagging for example, CRFs use the concatenation of all the different features instantiated from each template and the real-value of each feature is the occurred frequency. Actually, the subset of features $\kappa_j(\mathcal{D})$ applied by the $j$th template can be naturally formulated as a group. Each group of features stands for some specific meaning of the applied template, such as word, bi-gram, distance, orientation, and position related to the current token. Moreover, templates can be either defined arbitrarily by someone without any prior knowledge, or  designed specially by domain experts. It is easy to see that hundreds of templates are required for typical NLP tasks. Due to the diversity of features, the importance of each template should be assigned differently. However, due to numerous predefined templates, it is infeasible and tedious for users to select or weight the informative templates for the optimal prediction performance.

It is imperative to have a unified way to properly interpret and manage the diverse templates. In particular, a proper template weighting scheme can prune away poorly designed or conflicting templates, and amplify the effective templates, thereafter templates can be designed without cautions.
In the next subsections, we propose a novel structured prediction model with group sparsity, where features generated from the same template  naturally form a group, so as to interpret the importance of templates. The proposed model can be essentially deemed as a special MKL problem, where the base kernels are defined in accord with templates.
Then we solve this MKL in the primal by an efficient cutting plane algorithm.

\subsection{Model Formulation}

Given a training dataset $\mathcal{D}$, the $i^{th}$ input-output pair $(\mathcal{X}_i, \mathcal{Y}_i)$ can be represented by $\Phi(\mathcal{X}_i,\mathcal{Y}_i)=[\Phi_1(\mathcal{X}_i, \mathcal{Y}_i);~ \ldots;~ $ $\Phi_m(\mathcal{X}_i, \mathcal{Y}_i)]$ using the group representation of $\kappa(\mathcal{D})$ where semicolon is used to concatenate column vectors, for concise representation. The goal of structured prediction learning is to learn the hypotheses $f: \mathbb{X} \rightarrow \mathbb{Y}$.
According to Structural SVMs, the compatibility function $F:\mathbb{X} \times \mathbb{Y} \rightarrow \mathbb{R}$ over the input-output pairs are pursued and the prediction function can be derived by maximizing $F$ over the output space $\mathbb{Y}$ for a given input $\mathcal{X} \in \mathbb{X}$. The general hypotheses $f$ are the parameterized functions with parameter vector $\textbf{w}$ as
\begin{equation}
f(\mathcal{X};\textbf{w}) \!= \arg \max_{\mathcal{Y} \in \mathbb{Y} } F(\mathcal{X},\mathcal{Y};\textbf{w}) \!= \arg \max_{\mathcal{Y} \in \mathbb{Y} } \textbf{w}^T \Phi(\mathcal{X},\mathcal{Y}). \label{op:prediction}
\end{equation}
For structural outputs, the standard zero-one cost function frequently used in classification is not appropriate. Most applications need task-specific cost function, so we define a general cost function  $\Delta(\mathcal{Y},\mathcal{Y}')$ if the instance with the true output $\mathcal{Y}$ is assigned to be $\mathcal{Y}'$ with the property $\Delta(\mathcal{Y},\mathcal{Y})=0$. The margin re-scaling with general loss functions and linear penalty term can be formulated as a minimization of the regularized empirical risk:
\begin{small}
\begin{eqnarray}
\min_{\textbf{w},\boldsymbol{\xi} \geq 0} && \frac{1}{2} ||\textbf{w}||^2 + \frac{C}{n} \sum_{i=1}^n \xi_i \label{op:margin-scaling}\\
\textrm{s.t.} && \forall i, \forall \mathcal{Y}_i' \in \mathbb{Y}: \textbf{w}^T \delta \Phi^i(\mathcal{Y}_i') \geq \Delta(\mathcal{Y}_i,\mathcal{Y}_i') - \xi_i \nonumber
\end{eqnarray}
\end{small}\noindent
where $C$ is a trade-off parameter between training error minimization and margin maximization, and the difference of feature vectors is denoted by $\delta \Phi^i(\mathcal{Y}'_i)= \Phi(\mathcal{X}_i,\mathcal{Y}_i) - \Phi(\mathcal{X}_i,\mathcal{Y}_i')$.

As mentioned in Section \ref{sec:template}, group sparsity could be applied to this problem in terms of the naturally formed groups of features. According to Support Kernel Machines (SKMs) \cite{Bach04}, Problem (\ref{op:margin-scaling}) can be readily formulated with group feature representation $\Phi(\mathcal{X}_i, \mathcal{Y}_i)=[\Phi_1(\mathcal{X}_i, \mathcal{Y}_i);\ldots;$ $\Phi_m(\mathcal{X}_i, \mathcal{Y}_i)]$ as,
\begin{small}
\begin{eqnarray}
\min_{\textbf{w},\boldsymbol{\xi} \geq 0} && \Omega(\textbf{w}) + \frac{C}{n} \sum_{i=1}^n \xi_i \label{op:margin-scaling-mkl}\\
\textrm{s.t.} && \forall i, \forall \mathcal{Y}_i' \in \mathbb{Y}: \sum_{j=1}^m \textbf{w}_j^T \delta \Phi^i_j(\mathcal{Y}_i') \geq \Delta(\mathcal{Y}_i,\mathcal{Y}_i') - \xi_i, \nonumber
\end{eqnarray}
\end{small}\noindent
where $\textbf{w} = [\textbf{w}_1;\ldots;\textbf{w}_m]$, and the regularizer $\Omega(\textbf{w})$ can be defined as $\Omega(\textbf{w})=\frac{1}{2} \big( \sum_{j=1}^m ||\textbf{w}_j||\big)^2$ for SKM~\cite{Bach04}, $\Omega(\textbf{w})=\frac{1}{2}  \sum_{j=1}^m ||\textbf{w}_j||$ for group lasso~\cite{Bach08}, $\Omega(\textbf{w})=\frac{1}{2} \big( \sum_{j=1}^m ||\textbf{w}_j||^p\big)^{2/p}$ or $\Omega(\textbf{w})=\frac{1}{2} \big( \sum_{j=1}^m ||\textbf{w}_j||^p\big)^{2/p}+\frac{1}{2} \big( \sum_{j=1}^m ||\textbf{w}_j||\big)^2$ where $p\geq 1$ for generalized MKL~\cite{Kloft2010}. The number of constraints in Problem (\ref{op:margin-scaling-mkl}) depends on the specific structure in the space of $\mathbb{Y}$. In this paper, we mainly focus on sequence or tree structure which usually induces exponential number of constraints. Therefore, the general SKMs, group lasso algorithms, as well as the state-of-the-art MKLs, cannot be applied here. Recall that most algorithms focus on the dual problem of MKLs, but we propose an efficient cutting plane algorithm to solve Problem (\ref{op:margin-scaling-mkl}) in the primal form.

\subsection{Multiple Kernel Learning Trained in the Primal}

Problem (\ref{op:margin-scaling-mkl}) has exponential number of constraints, so the mixed norm regularizer $\Omega(\textbf{w})$ makes it even harder to solve. The following theorem states that Problem (\ref{op:margin-scaling-mkl}) can be equivalently formulated as 1-slack formulation \cite{Joachims09}.
\begin{theorem}
Problem (\ref{op:margin-scaling-mkl}) is equivalent to the problem
\begin{small}
\begin{eqnarray}
\min_{\textbf{w},\xi \geq 0} && \Omega(\textbf{w}) + C \xi \label{op:margin-scaling-mkl-one-slack}\\
\textrm{s.t.} && \forall [\mathcal{Y}_1',\ldots,\mathcal{Y}_n'] \in \mathbb{Y}^n: \nonumber \\
&& \frac{1}{n} \sum_{i=1}^n \sum_{j=1}^m \textbf{w}_j^T \delta \Phi^i_j(\mathcal{Y}_i') \geq \frac{1}{n} \sum_{i=1}^n \Delta(\mathcal{Y}_i,\mathcal{Y}_i') - \xi. \nonumber
\end{eqnarray}
\end{small}\noindent
\end{theorem}
\begin{proof}
The empirical risk can be derived as follows,
\begin{small}
\begin{eqnarray*}
\frac{1}{n} \sum_{i=1}^n \xi_i \!\!\!\!&=&\!\!\!\! \frac{1}{n} \sum_{i=1}^n \max_{\mathcal{Y}_i' \in \mathbb{Y}} \bigg[\Delta(\mathcal{Y}_i,\mathcal{Y}_i') - \sum_{j=1}^m \textbf{w}_j^T \delta \Phi^i_j(\mathcal{Y}_i')\bigg] \\
\!\!\!\!&=&\!\!\!\! \max_{[\mathcal{Y}_1',\ldots,\mathcal{Y}_n'] \in \mathbb{Y}^n} \frac{1}{n}\sum_{i=1}^n\bigg[\Delta(\mathcal{Y}_i,\mathcal{Y}_i') - \sum_{j=1}^m \textbf{w}_j^T \delta \Phi^i_j(\mathcal{Y}_i')\bigg] \!= \!\xi
\end{eqnarray*}
\end{small}\noindent
where the second equality holds since for any given $\mathbf{w}$ each $\mathcal{Y}_i'$ can be optimized independently due to the linearly decomposed property in $\mathcal{Y}_i'$ ~\cite{Joachims09}. This completes the proof.
\end{proof}

Hereby, we propose Algorithm \ref{algo:smkl} to solve Problem (\ref{op:margin-scaling-mkl-one-slack}). This algorithm constructs a working set $\mathcal{W}$ iteratively. In each iteration, the most violated constraint is found, and then added into the working set $\mathcal{W}$. After that, a subproblem is solved over $\mathcal{W}$ in order to obtain a new solution $\textbf{w}$.

Assume that the most violated outputs $(\widehat{\mathcal{Y}}_1,\ldots,\widehat{\mathcal{Y}}_n)$ can be accessed given  $\textbf{w}$. After $s$ iterations, we can construct a set of most violated labels $\mathcal{W}^s$,   the subproblem in Algorithm \ref{algo:smkl} is formulated as
\begin{small}
\begin{eqnarray}
\min_{\textbf{w},\xi \geq 0} && \Omega(\textbf{w}) + C \xi \label{op:margin-scaling-mkl-one-slack-subproblem}\\
\textrm{s.t.} && \xi \geq q^r + \sum_{j=1}^m \textbf{w}_j^T \textbf{p}_j^r, \forall r=1,\ldots,s \nonumber
\end{eqnarray}
\end{small}\noindent
where $\textbf{p}_j^r = -\frac{1}{n} \sum_{i=1}^n \delta \Phi^i_j(\widehat{\mathcal{Y}}_i^r)$ and $q^r = \frac{1}{n} \sum_{i=1}^n \Delta(\mathcal{Y}_i,\widehat{\mathcal{Y}}_i^r)$. Note that subproblem (\ref{op:margin-scaling-mkl-one-slack-subproblem}) has $s$ constraints.
For sake of simplicity, we use $\Omega(\textbf{w})=\frac{1}{2} \big( \sum_{j=1}^m \|\textbf{w}_j\|\big)^2$ as the showcase in this paper~\footnote{The dual of other regularizers can be found in~\cite{Kloft2010}.},  the conic dual \cite{Bach04} of Problem (\ref{op:margin-scaling-mkl-one-slack-subproblem}) can be readily derived as
\begin{small}
\begin{eqnarray}
\max_{\alpha \in \mathcal{A}_s} \max_{\theta} && - \theta + \sum_{r=1}^s \alpha_r q^r \label{op:qcqp-dual}\\
\textrm{s.t.} && \frac{1}{2} \alpha^T Q^j \alpha \leq \theta, \forall j=1,\ldots,m \nonumber
\end{eqnarray}
\end{small}\noindent
where $\mathcal{A}_s = \{ \sum_{r=1}^s \alpha_r \leq C, \alpha_r \geq 0, \forall r=1,\ldots,s\}$, and $Q_{r,r'}^j = \langle \textbf{p}^r_j, \textbf{p}^{r'}_j \rangle$. For completeness, we give the derivation in Appendix \ref{proof:qcqp}. Problem (\ref{op:qcqp-dual}) is in form of a quadratically constrained quadratic programming (QCQP) problem, which is similar to the multiple kernel learning problem with a small size of constraints. Furthermore, the primal solutions can be recovered by
\begin{small}
\begin{equation}
\textbf{w}_j = -\mu_j \sum_{r=1}^s \alpha_r \textbf{p}_j^r, \forall j=1,\ldots,m, \label{eq:recover-w}
\end{equation}
\end{small}\noindent
where $\mu$ is the Lagrangian multipliers, each of which corresponds to one constraint in (\ref{op:qcqp-dual}). Together with the sparsity of $\alpha$ and the sparsity of $\mu$, the primal variable $\mathbf{w}$ is sparse and so the recovery of $\mathbf{w}$ is also fast. The algorithm stops if no constraint is found with the desired precision or the maximum number of iterations is reached.

\begin{algorithm}
\begin{small}
   \caption{ Multiple Template Learning }
   \label{algo:smkl}
\begin{algorithmic}[1]
    \STATE \textbf{Input}: $\mathcal{D} = \{ (\mathcal{X}_1, \mathcal{Y}_1),\ldots,(\mathcal{X}_n, \mathcal{Y}_n) \}, C, \epsilon$
    \STATE $\textbf{w}=\textbf{0}$, $\mathcal{W} = \emptyset$,
    \REPEAT
       \STATE Find the most violated $(\widehat{\mathcal{Y}}_1,\ldots,\widehat{\mathcal{Y}}_n)$
       \STATE $\mathcal{W} := \mathcal{W} \cup \{(\widehat{\mathcal{Y}}_1,\ldots,\widehat{\mathcal{Y}}_n) \}$
       \STATE Obtain $\textbf{w}$ by solving subproblem (\ref{op:qcqp-dual}) over $\mathcal{W}$ according to (\ref{eq:recover-w})
    \UNTIL{$\epsilon$-optimal}
    \STATE \textbf{return} $\textbf{w}$
\end{algorithmic}
\end{small}
\end{algorithm}

As will be shown in Section~\ref{sec:converge}, when $\Omega(\textbf{w})=\frac{1}{2} \big( \sum_{j=1}^m \|\textbf{w}_j\|\big)^2$, Algorithm~\ref{algo:smkl} can converge to an $\epsilon$-optimality in a finite number of iterations. Empirically, the number of iteration $s$ needed for Algorithm \ref{algo:smkl} to reach $\epsilon$-optimal convergence is very small.  Therefore, the QCQP problem in (\ref{op:qcqp-dual}) with $s+1$ variables and $m+1$ constraints can be solved efficiently by a QCQP toolbox, such as Mosek \cite{Andersen02}. Since Mosek simultaneously solves the primal and its dual form, the weights $\mu$ for each group of features can be obtained at the same time. Alternatively, one can apply other efficient MKL algorithms \cite{Xu09,Rakotomamonjy08,Sonnenburg06} to solve (\ref{op:qcqp-dual}) when a sophisticated QCQP solver is not available, but the solution may not be as accurate as that of Mosek.

It is worth mentioning that the proposed method obtains $\textbf{w}_j = -\mu_j \sum_{r=1}^s \alpha_r \textbf{p}_j^r, \forall j=1,\ldots,m$; while Structural SVMs do not consider the weights $\mu$, \emph{ie.} a uniform weighting for templates.
 Another good property is that,  when $\Omega(\textbf{w})=\frac{1}{2} \big( \sum_{j=1}^m ||\textbf{w}_j||\big)^2$,  the sparsity of MTL$^{struct}$ is not only on $\alpha$ but also on $\mu$ due to the feasible domain $\mathcal{A}_s$ and quadratic constraints in Problem (\ref{op:qcqp-dual}), but the cutting plane algorithm for the uniform weighting does not enforce the sparsity on the feature groups.

Finding the most violated constraint with outputs $(\widehat{\mathcal{Y}}_1,\ldots,\widehat{\mathcal{Y}}_n)$ is to solve the following problem
\begin{small}
\begin{equation}
\arg \max_{ (\mathcal{Y}_1',\ldots,\mathcal{Y}_n')  \in \mathbb{Y}^n} \sum_{i=1}^n \Delta(\mathcal{Y}_i,\mathcal{Y}_i') - \sum_{i=1}^n \sum_{j=1}^m \textbf{w}_j^T \delta \Phi^i_j(\mathcal{Y}_i'), \label{op:m-inference}
\end{equation}
\end{small}\noindent
where the definition of cost $\Delta(\mathcal{Y}_i,\mathcal{Y}_i')$ and the feature mapping $\Phi$ depend on the specific tasks. The input-output pairs in $\mathcal{D}$ are generally considered i.i.d., so Problem (\ref{op:m-inference}) can be generally decomposed into $n$ independent optimization problems as
\begin{small}
\begin{equation}
\widehat{\mathcal{Y}}_i = \arg \max_{\mathcal{Y}_i' \in \mathbb{Y}} \Delta(\mathcal{Y}_i,\mathcal{Y}_i') - \sum_{j=1}^m \textbf{w}_j^T \delta \Phi^i_j(\mathcal{Y}_i'), \forall i=1,\ldots,n. \label{op:1-inference}
\end{equation}
\end{small}\noindent

In the following subsections, we present the convergence analysis and complexity analysis of Algorithm \ref{algo:smkl}. After that, we deal with specific tasks, namely sequence labeling and dependency parsing. Once the parameter $\textbf{w}$ is learned, we can do prediction by solving an inference problem in (\ref{op:prediction}) given an input $\mathcal{X} \in \mathbb{X}$.

\subsection{Convergence Analysis} \label{sec:converge}

The termination condition of Algorithm \ref{algo:smkl} is defined as $ R_{emp}(\textbf{w}_s) - R_s(\textbf{w}_s) < \epsilon$ where the risk of upper bound and lower bound of Problem (\ref{op:margin-scaling-mkl-one-slack}) are
\begin{displaymath}
\begin{small}
\begin{array}{l}
R_{emp}(\textbf{w}) = \frac{1}{n} \sum_{i=1}^n \max_{\mathcal{Y}_i' \in \mathbb{Y}} \left( \Delta(\mathcal{Y}_i,\mathcal{Y}_i') - \sum_{j=1}^m \textbf{w}_j^T \delta \Phi^i_j(\mathcal{Y}_i') \right),\\
R_s(\textbf{w}) = \max_{r=1,\ldots,s} \left( \sum_{j=1}^m \textbf{w}_j^T \textbf{p}_j^r + q^r \right).
\end{array}
\end{small}
\end{displaymath}\noindent
The maximum number of iterations could also be used to terminate the algorithm considering the time and space limitations.

Notice that the convergence proof for bundle method or cutting plane algorithm in \cite{Teo10,Teo2007} does not apply in our case as the Fenchel dual of mixed norm fails to satisfy the strong convexity assumption if $m > 1$. As $m=1$, Algorithm~\ref{algo:smkl} is exactly the bundle method \cite{Teo10}. When $m > 1$, Theorem \ref{theorem:QCQP} shows the convergence of Algorithm~\ref{algo:smkl} which can converge to $\epsilon$-optimality in a finite number of iterations. The proof is shown in Appendix~\ref{app:proof-theorem2}.

\begin{theorem} \label{theorem:QCQP}
For any $0 < C, 0 < \epsilon \leq 4 R^2 C$ and any training example $(\mathcal{X}_1,\mathcal{Y}_1),\ldots,(\mathcal{X}_n,\mathcal{Y}_n)$, Algorithm \ref{algo:smkl} converges to the desired precision $\epsilon$ after at most,
\begin{small}
\begin{eqnarray*}
\left\lceil \log_2 \left( \frac{\Delta}{4 R^2 C} \right) \right\rceil + \left \lceil \frac{16 R^2 C}{\epsilon} \right\rceil
\end{eqnarray*}
\end{small}
iterations, where $R^2 = \max_{i,\mathcal{Y}'} \|\Phi(\mathcal{X}_i,\mathcal{Y}_i) - \Phi(\mathcal{X}_i,\mathcal{Y}_i')\|^2$, $\Delta = \max_{i,\mathcal{Y}'_i} \Delta(\mathcal{Y}_i',\mathcal{Y}_i)$ and $\lceil . \rceil$ is the integer ceiling function.
\end{theorem}

\subsection{Complexity Analysis} \label{sec:complexity}

In the following applications, we only use the linear kernel in the proposed algorithm. According to \cite{Lanckriet04}, given a QCQP problem with $s$ variables and $m$ quadratic constraints, the optimization tools such as Mosek can yield the worst-case complexity of $O(m s^3)$ by using interior-point method. At the $s^{th}$ iteration, Algorithm \ref{algo:smkl} takes at most $O(m s^3)$ time for solving the reduced QCQP problem (\ref{op:qcqp-dual}), $O(s^2)$ for computing $\xi$, $O(s |\kappa(\mathcal{D})| )$ for $\mathbf{w}$, $O(n)$ for $q^s$, $O(n |\kappa(\mathcal{D})|)$ for $\mathbf{p}^s$ and $O(s |\kappa_j(\mathcal{D})| )$ for adding a row/column to $Q^j$ since $Q^j$ can be computed incrementally, $\forall j=1,\ldots,m$.
Another major training cost is to find the most violated constraint. For sequence labeling task, suppose the average length of sequences are $l$ and the number of classes are $c$, then it takes $O(lc|\kappa(\mathcal{D})|)$ time to find the most violated constraint. Similarly, for dependency parsing task, it takes $O(l^3 |\kappa(\mathcal{D}|)$ or $O(l^2 |\kappa(\mathcal{D}|)$ time for projected and non-projected dependency tree structure, respectively. Hence, the overall time complexity is $O(m s^3 + s |\kappa(\mathcal{D})| + n u |\kappa(\mathcal{D})|)$, where $u=\{lc,l^2,l^3\}$ for different inference problems.

For sparse and high dimensional  data, Algorithm \ref{algo:smkl} can efficiently exploit the sparsity of the feature vectors, so that computing of inner product for $Q$ only depends on the number of non-zeros $z$. So the overall time complexity is  reduced to $O(m s^3 + s z + n u z)$.
 According to Theorem \ref{theorem:QCQP}, the algorithm converges to the $\epsilon$-optimal solution in $O(\frac{C}{\epsilon})$. And empirical observations show that several hundreds of iterations are enough for good performance. Moreover, we also employ the cut pruning strategy in the implementation, which has been used in Structural SVM \cite{Joachims09},  to speed up the learning process. So it is faster than the analysis in practice.
Comparing with Structural SVM with $l_2$ norm, except Structural SVM needs to solve a QP problem (takes $O(s^3)$ time) instead of solving a QCQP problem, the complexity for other parts are the same. So the overall time complexity of our proposed MTL$^{struct}$ is very comparable with Structural SVM.

 OnlineMKL \cite{Martins10} attempts to solve the similar problem by online learning where two operations are needed at each iteration: the gradient descent step and proximal step. By exploiting the sparsity of features, the gradient step takes $O(z)$ time to update $\mathbf{w}$, $O(uz)$ to find the most violated configuration for the randomly sampled instance, and $O(m z)$ for incremental update of $\|\mathbf{w}_j\|, \forall j=1,\ldots,m$. However, the proximal step needs to scale $\mathbf{w}$ according to these norms, so it has to take $O(|\kappa(\mathcal{D})|)$ time where $\mathbf{w}$ is hard to exploit the sparsity. This is different from Algorithm \ref{algo:smkl} with both sparsity on $\alpha$ and $\mu$. The overall time complexity for each epoch takes $O(n ( uz + mz + |\kappa(\mathcal{D})| ))$ time in OnlineMKL. Generally, it takes tens of epochs to gain a good performance. Hence, for high dimension but very sparse data, i.e. more than $10$ million, $O(n |\kappa(\mathcal{D})|)$ will dominate the time complexity for OnlineMKL, while Algorithm \ref{algo:smkl} is only dominated by
the inference cost ($O(nuz)$). Taking  the sequence labeling task for example, suppose that $20$ epochs for OnlineMKL, and $500$ iterations for Algorithm \ref{algo:smkl} used in the following experiments, the time complexity for OnlineMKL is $O(20 n |\kappa(\mathcal{D})|)$ while that of Algorithm \ref{algo:smkl} is $O(500 n lc z)$ only. OnlineMKL is slower than Algorithm \ref{algo:smkl} since $|\kappa(\mathcal{D})| \gg 25 lcz$ where $l$, $c$ and $z$ are usually very small (around tens). In the following applications, we will empirically demonstrate OnlineMKL is quite slow.

\section{Extension to $p$-Block-Norm Regularization} \label{sec:pmkl}

In this section, we consider the generalized form of mixed norm regularizer $\Omega(\textbf{w})$=$\frac{1}{2} \big( \sum_{j=1}^m \|\textbf{w}_j\|^p\big)^{2/p}$ with $p> 1$~\cite{Kloft2010}. According to Theorem~1 and (\ref{op:margin-scaling-mkl-one-slack-subproblem}), we obtain the following subproblem
\begin{small}
\begin{eqnarray}
\min_{\textbf{w},\xi \geq 0} && \frac{1}{2} \bigg( \sum_{j=1}^m \|\textbf{w}_j\|^p\bigg)^{2/p} + C \xi \label{op:margin-scaling-pmkl-one-slack-subproblem}\\
\textrm{s.t.} && \xi \geq q^r + \sum_{j=1}^m \textbf{w}_j^T \textbf{p}_j^r, \forall r=1,\ldots,s. \nonumber
\end{eqnarray}
\end{small}\noindent
after $s$ iterations in Algorithm \ref{algo:smkl}. The following proposition gives the dual problem of (\ref{op:margin-scaling-pmkl-one-slack-subproblem}).
\begin{proposition} \label{prop:pmkl}
The dual problem of (\ref{op:margin-scaling-pmkl-one-slack-subproblem}) with $p > 1$ is
\begin{small}
\begin{eqnarray}
\max_{\alpha \in \mathcal{A}_s} - \frac{1}{2} \bigg( \sum_{j=1}^m \left(\sqrt{\alpha^T Q^j \alpha} \right)^{p^*}  \bigg)^{2/p^*}  + \sum_{r=1}^s \alpha_r q^r, \label{op:pmkl-dual}
\end{eqnarray}
\end{small}\noindent
where $p^* = \frac{p}{p-1}$. The weights of kernels are recovered as
\begin{small}
\begin{eqnarray}
\mu_j = \bigg( \sum_{j=1}^m (\sqrt{\alpha^T Q^j \alpha})^{\frac{p}{p-1}} \bigg)^{\frac{p-2}{p}}  \left( \sqrt{\alpha^T Q^j \alpha} \right)^{\frac{2-p}{p-1}},
\end{eqnarray}
\end{small}\noindent
$\forall j=1,\ldots,m$,
and primal variables can be recovered by
\begin{small}
\begin{equation}
\textbf{w}_j = -\mu_j \sum_{r=1}^s \alpha_r \textbf{p}_j^r, \forall j=1,\ldots,m.
\end{equation}
\end{small}
\end{proposition}
The detailed proof of Proposition \ref{prop:pmkl} is shown in Appendix \ref{proof:pmkl}.

Problem (\ref{op:pmkl-dual}) is a smooth function with $p>1$ and the feasible domain of $\alpha$ in a simplex $\mathcal{A}_s$. We can employ projected gradient descent to solve (\ref{op:pmkl-dual}) where the projection step can be readily solved by efficient projections onto simplex method~\cite{Duchi2008}. Since the inference process is the same as Algorithm \ref{algo:smkl}, the only difference is to solve the subproblem (\ref{op:pmkl-dual}) instead of (\ref{op:qcqp-dual}). Hence, Algorithm \ref{algo:smkl} can be easily extended to solve MTL$^{struct}$ with $p$-block norm regularization.

\section{Experiments} \label{sec:experiments}

As aforementioned, templates are employed widely in various applications. To verify the proposed method, we explore it on three tasks: Chinese Word Segmentation \cite{Xue03,Peng04}, Named Entity Recognition \cite{McCallum03}, and Dependency Parsing \cite{McDonald05a}. The first two tasks are popularly transformed into the structured prediction problem with the output as a sequence structure \cite{Peng04,McCallum03}, while dependency parsing is cast as the structured prediction problem with the output as a spanning tree \cite{McDonald05a}. In what follows, we will first give the general setting of these tasks, and then the task-specific settings and their experimental results are shown separately.

\begin{table}
    \caption{Benchmark datasets for Chinese word segmentation.}
    \label{tab:seg_data_desciption}
    \begin{center}
    \begin{tabular}{l|r|r|r|r}
    \hline
    \multirow{2}{*}{Dataset} & \multicolumn{2}{|c|}{\# Sentence}  & \multicolumn{2}{|c}{\#Features} \\ \cline{2-5}
         & Train~  & Test~                   & TP1~~~ & TP2~~~~ \\
    \hline
    AS      & 708,953 & 14,429 & 31,587,879 &53,599,827\\
    MSR     & 86,924  & 3,985 & 17,055,615&31,228,275\\
    CityU   &  53,019 & 1,492 &  13,196,613 &25,143,720 \\
    PKU     & 19,054  & 1,944 & 9,605,457 &18,717,687 \\
    \hline
    \end{tabular}
    \end{center}
\end{table}

\subsection{Experimental Setting}

\begin{table}
\caption{Two set of templates used for Chinese word segmentation task: TP1 and TP2. The symbol '-' means that this template is not used in the setting. } \label{tab:seg_templates}
\begin{center}
\begin{tabular}{l|c|c|l|l}
\hline
\multirow{2}{*}{Type} & \multicolumn{2}{|c|}{Index} & \multirow{2}{*}{Name} & \multirow{2}{*}{Template}\\ \cline{2-3}
     & TP1 & TP2                   &     & \\
\hline
uni-gram    &1&1&U00&\%x[-2,0] \\
    &2&2&U01&\%x[-1,0]\\
    &3&3&U02&\%x[0,0] \\
    &4&4&U03&\%x[1,0] \\
    &5&5&U04&\%x[2,0] \\
\hline
bi-gram &9&1&   U08&\%x[-1,0]/\%x[0,0] \\
    &10&10&U09&\%x[0,0]/\%x[1,0]\\
    & -&11&U10&\%x[-2,0]/\%x[-1,0] \\
    & -&12&U11&\%x[1,0]/\%x[2,0] \\
    & -&13&U12&\%x[-2,0]/\%x[0,0] \\
    & -&14&U13&\%x[-1,0]/\%x[1,0] \\
    & -&15&U14&\%x[0,0]/\%x[2,0] \\
    & -&16&U15&\%x[-2,0]/\%x[1,0] \\
    & -&17&U16&\%x[-1,0]/\%x[2,0] \\
\hline
tri-gram &6&6&   U05&\%x[-2,0]/\%x[-1,0]/\%x[0,0]\\
    &7&7&U06&\%x[-1,0]/\%x[0,0]/\%x[1,0] \\
    &8&8&U07&\%x[0,0]/\%x[1,0]/\%x[2,0] \\
\hline
    & 11 & 18& B & state transition \\
\hline
\end{tabular}
\end{center}
\end{table}

\begin{table}[t]
\caption{The performance (TP1/TP2) of Chinese word segmentation for comparing methods over five randomly sampled training and testing datasets. For each method, the first row is the mean and the second row is the corresponding standard deviation. The best results across the compared methods are shown in bold on TP1 and TP2, individually.} \label{tab:five-random-results}
\scriptsize
\begin{center}
\begin{tabular}{c|l|cccc}
\hline
Dataset & Algorithm       & Recall      & Prec          & F1       & $\textrm{R}_{iv}$ \\
\hline\hline
\multirow{12}{*}{PKU}& \multirow{2}{*}{CRF ($\ell_2$)}   &96.4 / 96.0  & \textbf{96.7} / 96.0 & 96.6 / 96.0 & 97.2 / 96.8 \\
                     &                                   &0.10 / 0.10  & 0.10 / 0.10  & 0.12 / 0.10  &0.06 / 0.06 \\
\cline{2-6}
& \multirow{2}{*}{CRF ($\ell_1$)}                        &95.7 / 95.5  & 95.9 / 95.3 & 95.8 / 95.4 & 96.4 / 96.3 \\
&                                                        &0.10 / 0.06  & 0.15 / 0.10 & 0.10 / 0.06  & 0.06 / 0.06 \\
\cline{2-6}
& \multirow{2}{*}{IRW-CRF}                              & \textbf{97.3} / 95.9  & 96.5 / 95.8 & 96.4 / 95.8 & 97.1 / 96.7 \\
&                                                       & 0.15 / 0.00 &  0.20 / 0.06  &  0.15 / 0.06&  0.06 / 0.06\\
\cline{2-6}
& \multirow{2}{*}{$\textrm{SVM}^{hmm}$}                 &96.6 / 96.5 & \textbf{96.7} / 96.5 & 96.6 / 96.5 & 97.4 / 97.2\\
&                                                       &0.12 / 0.10 & 0.12 / 0.06 & 0.10 / 0.06& 0.12 / 0.06\\
\cline{2-6}
& \multirow{2}{*}{OnlineMKL}                            &95.9 / 96.2 & 95.9 / 96.3 & 95.9 / 96.3 &96.6 / 97.0\\
&                                                       &0.06 / 0.15 &0.20 / 0.15 &0.10 / 0.15 &0.06 / 0.17\\
\cline{2-6}
& \multirow{2}{*}{$\textrm{MTL}^{hmm}$}                 & 97.1 / \textbf{97.2} & \textbf{96.7} / \textbf{96.8}  & \textbf{96.9} / \textbf{97.0}  & \textbf{98.0} / \textbf{98.1} \\
&                                                       &0.10 / 0.10 & 0.10 / 0.06 & 0.10 / 0.10 & 0.06 / 0.00\\
\hline
\hline
\multirow{8}{*}{CityU}& \multirow{2}{*}{CRF ($\ell_2$)} & 96.3 / 95.9  & 96.6 / 95.8 & 96.4 / 95.9 & 96.9 / 96.6 \\
                     &                                  & 0.06 / 0.06  & 0.06 / 0.06 & 0.06 / 0.10 & 0.06 / 0.06 \\
\cline{2-6}
& \multirow{2}{*}{CRF ($\ell_1$)}                       & 97.0 / 96.5  & 97.2 / 96.5 & 97.1 / 96.5 & 97.6 / 97.1 \\
&                                                       & 0.06 / 0.06  & 0.06 / 0.00 &  0.06 / 0.00  & 0.06 / 0.00 \\
\cline{2-6}
& \multirow{2}{*}{IRW-CRF}                              & 97.0 / 96.4  &  97.2 / 96.4 &  97.1 / 96.4 & 97.6 / 97.0 \\
&                                                       & 0.12 / 0.12 &  0.10 / 0.10  & 0.12 / 0.10 &  0.12 / 0.06\\
\cline{2-6}
& \multirow{2}{*}{$\textrm{SVM}^{hmm}$}                 & 97.0 / 96.9 & 97.1 / 96.8 & 97.0 / 96.8 & 97.6 / 97.4\\
&                                                       & 0.00 / 0.12 & 0.06 / 0.00 & 0.06 / 0.06& 0.06 / 0.06\\
\cline{2-6}
& \multirow{2}{*}{$\textrm{MTL}^{hmm}$}                 & \textbf{97.6} / \textbf{97.7}& \textbf{97.4} / \textbf{97.5}  & \textbf{97.5} / \textbf{97.6} & \textbf{98.4} / \textbf{98.4} \\
&                                                       & 0.06/ 0.00 & 0.00 / 0.00 & 0.06 / 0.00 & 0.06 / 0.06 \\
\hline
\end{tabular}
\end{center}
\end{table}

\begin{figure}
\centering
\begin{tabular}{cc}
\includegraphics[width=0.5\textwidth]{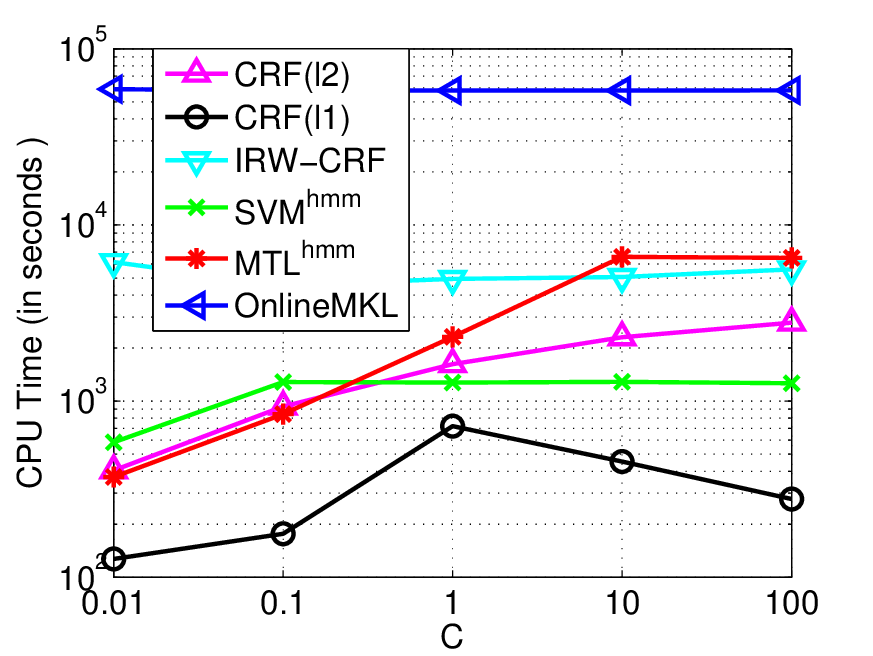}&\!\!\!\!\includegraphics[width=0.5\textwidth]{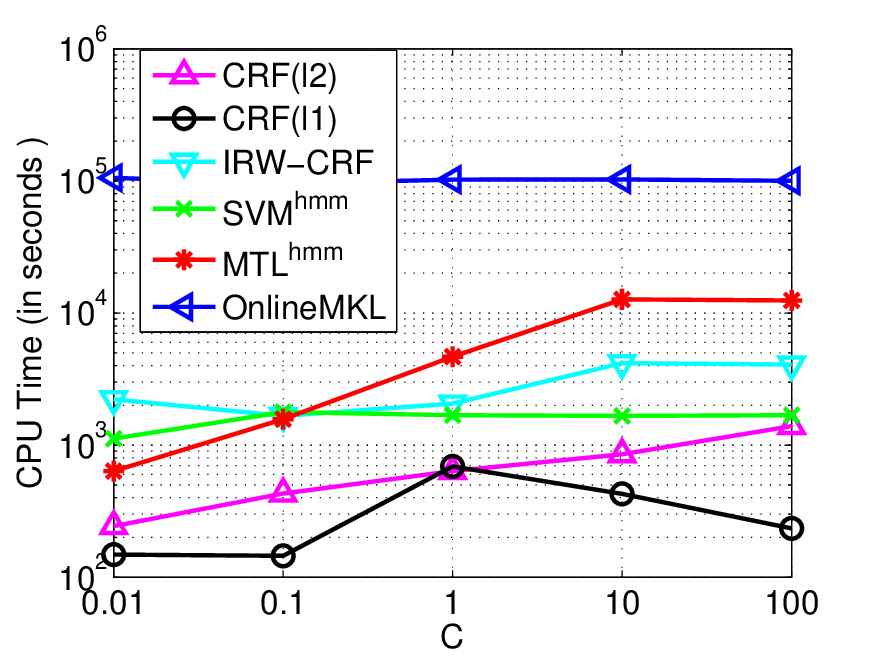}\\
(a) TP1 & (b) TP2 \\
\end{tabular}
\caption{The average CPU time on PKU dataset by varying $C$ on five randomly sample of training and testing datasets.}
\label{fig:time-pku}
\end{figure}

In the sequence learning, high-order dependence among labels and state-state transition with observation in the linear chain could be considered, but the state-state transition feature without observations are considered in this paper for two reasons. First, the inference problem (\ref{op:m-inference}) can be solved efficiently and effectively by the Viterbi algorithm \cite{Viterbi67} for Hamming loss \cite{Tsochantaridis05}. Second, there exist toolboxes with state-of-the-art methods for the fair comparison. The methods for comparison are listed as follows:

\begin{itemize}
\item \textbf{CRF}($\ell_2$) (\textbf{CRF}($\ell_1$)) \cite{Lafferty01}: CRF with a Gaussian (Laplacian) prior defined by CRF++ toolbox\footnote{http://crfpp.sourceforge.net/} in terms of the command option ``-a''.

\item \textbf{IRW-CRF}: Solving weighted CRF($\ell_2$) by iterative re-weighted least square method with  the update rule for weights as $(||\mathbf{w}_j||^2+\epsilon)^{-1}$~\cite{Wipf2010} and warm start using \textbf{CRF}($\ell_2$) solution. To avoid numerical instabilities, we fixed $\epsilon=0.001$ and the maximum iteration as $10$.

\item \textbf{SVM}$^{hmm}$ \cite{Tsochantaridis05}: Structural SVM for sequence learning by $\textrm{SVM}^{hmm}$ toolbox\footnote{http://www.cs.cornell.edu/People/tj/svm\_light/svm\_hmm.html}. The recommended termination condition ``-e 0.5'' is used in following experiments.

\item \textbf{OnlineMKL} \cite{Martins10}: Online MKL for structured prediction with C++ implementation which is provided by the authors. The number of training epochs is set to $20$ following \cite{Martins10}\footnote{In the following experiments, we will show that the large number of epochs is infeasible for the large scale high-dimensional data.}.
\item \textbf{MTL}$^{hmm}$: The instantiated method of the proposed MTL$^{struct}$ framework for sequence learning problem is implemented in C++. The termination condition is the same as \textbf{SVM}$^{hmm}$.
\end{itemize}

\begin{table}
\caption{The Chinese word segmentation results (TP1/TP2) of comparing methods on all the testing datasets. The best results across the compared methods are shown in bold on TP1 and TP2, individually. }
\label{tab:seg}
\scriptsize
\begin{center}
\begin{tabular}{c|l|cccc}
\hline
Dataset & Method       & Recall      & Prec          & F1       & $\textrm{R}_{iv}$ \\
\hline\hline
\multirow{4}{*}{AS}& CRF ($\ell_2$)          & \textbf{95.6} / 95.1 &   \textbf{94.4} / 93.8 &   95.0 / 94.4   &     97.0 / 96.4    \\
& CRF ($\ell_1$)          & 95.4 / 95.0  &   94.1 / 93.4 &   94.8 / 94.2   &     96.7 / 96.4    \\
& IRW-CRF      &  \textbf{95.6} / 95.1      & 94.3 / 93.8 & \textbf{94.9} / 94.5 & 97.0 / 96.5  \\
& $\textrm{SVM}^{hmm}$    & 94.6 / 94.3 &   93.2 / 93.3 &   93.9 / 93.8 &     96.0 / 95.6 \\
& $\textrm{MTL}^{hmm}$    & \textbf{95.6} / \textbf{95.8} &   93.9 / \textbf{93.9} &   94.7 / \textbf{94.8} &     \textbf{97.2} /\textbf{97.3}   \\
\hline \hline
\multirow{4}{*}{MSR}& CRF ($\ell_2$)          & 96.5 / 95.7&    {96.7} / 96.0&    96.6 / 95.8&    97.3 / 96.5  \\
& CRF ($\ell_1$)          & 96.2 / 95.7 &	96.2 / 95.6 &	96.2 / 95.6	& 	96.9 / 96.5 \\
& IRW-CRF  & 96.6 / 95.8  & \textbf{96.8} / 96.0 & \textbf{96.7} / 95.9 & 97.5 / 96.6\\
& $\textrm{SVM}^{hmm}$    & 96.7 / 96.1&    96.5 / \textbf{96.5}&    96.6 / 96.3&     97.6 / 96.7 \\
& $\textrm{MTL}^{hmm}$    & \textbf{96.9} / \textbf{97.0}&    96.5 / 96.4&    \textbf{96.7} / \textbf{96.7} &   \textbf{97.9} / \textbf{98.0}\\
\hline \hline
\multirow{4}{*}{CityU}& CRF ($\ell_2$)            & 94.1 / 92.8 & {94.3} / 92.6&    94.2 / 92.7&       96.3 / 95.1 \\
& CRF ($\ell_1$)            & 93.4 / 92.9 &	93.6 / 92.9 &	93.5 / 92.9 &		95.6 / 95.1 \\
& IRW-CRF  & 94.1 / 93.5  & \textbf{94.3} / 93.3 & 94.2 / 93.4 & 96.3 / 95.8\\
& $\textrm{SVM}^{hmm}$ & 94.3 / 94.1&  94.1 / 93.5&    94.2 / 93.8&       96.4 / 96.4 \\
& $\textrm{MTL}^{hmm}$      & \textbf{95.2} / \textbf{95.2}&  94.2 / \textbf{94.3}&    \textbf{94.7} / \textbf{94.8}&    \textbf{97.6} / \textbf{97.6} \\
\hline\hline
\multirow{4}{*}{PKU}& CRF ($\ell_2$)         & 92.7 / 92.2&    \textbf{94.2} / 93.3&    93.4 / 92.7&        94.7 / 94.2   \\
& CRF ($\ell_1$)         & 91.6 / 91.8 &	93.1 / 92.4 &	92.4 / 91.7 &		93.8 / 93.4 \\
& IRW-CRF  & 92.7 / 92.4  & 94.1 / 93.3  & 93.4 / 92.9 & 94.8 / 94.3\\
& $\textrm{SVM}^{hmm}$& 92.6 / 93.0&    93.7 / 93.4&    93.2 / 93.2&       94.9 / 95.0   \\
& OnlineMKL    & 89.9 / 91.0  &  89.2 / 90.5 &    89.6 / 90.7 &      91.7 / 92.7    \\
& $\textrm{MTL}^{hmm}$    & \textbf{93.8} / \textbf{94.0}&    93.3 / \textbf{93.6}&    \textbf{93.6} / \textbf{93.8}&      \textbf{96.1} / \textbf{96.2}    \\
\hline
\end{tabular}
\end{center}
\end{table}

All above comparing methods have the tradeoff parameter, so we rescale them into the grid of $\{10^{-2}, 10^{-1},10^0, 10^1, 10^2\}$ such that the objectives of all methods are scaled with a constant. In the following experiments, we set the $C$ of MTL$^{hmm}$ in the grid of $n \times \{10^{-2}, 10^{-1},10^0, 10^1, 10^2\}$, while the parameters of other methods are adjusted accordingly.

In the dependency parsing, we follow the structured prediction framework with the loss defined as the number of words that have the incorrect parent \cite{McDonald05a}. According to \cite{McDonald05}, there are two algorithms which can solve Problem (\ref{op:m-inference}) in terms of two types of dependency trees, respectively. For projective dependency tree, Eisner algorithm \cite{Eisner96} has a runtime of $O(l^3)$, while Chu-Liu-Edmonds \cite{Chu65} provides non-projective parsing complexity $O(l^2)$ where $l$ is the length of the input sentence.
Non-projective case is adopted according to the property of datasets used in the following experiments.
For fair comparison, we mainly focus on comparing the following methods:
\begin{itemize}
\item \textbf{MSTParser} \cite{McDonald05a}:  Online large margin method for dependency parsing with the available toolbox\footnote{http://sourceforge.net/projects/mstparser/}. We vary the number of epochs in the set of $\{20,30,40\}$.
\item \textbf{MTL}$^{parse}$: The instantiated method of the proposed MTL$^{struct}$ framework for dependency parsing.
\end{itemize}
The setting of MTL$^{parse}$ is similar to the task of sequence learning, but with a fixed $C$=$n \times 10^2$. We notice that more epochs does not affect the performance of MSTParser much.

The last experiments are conducted for the analysis of MTL$^{struct}$ with $p$-block-norm regularization ($p$\textbf{MTL}$^{struct}$) in the setting of sequence labeling problems ($p$\textbf{MTL}$^{hmm}$). We adopt the same setting as that of \textbf{MTL}$^{hmm}$, as well as vary the parameter $p$ in the set of $\{ 4/3, 2, 4, \infty \}$ ~\cite{Kloft2010} to compare with \textbf{MTL}$^{hmm}$ such that we can obtain the influence of the $p$-block-norm regularization in the sequence labeling tasks.

For fair comparisons, all comparing methods have the same set of templates(features) as the input, and the best results of each method are reported on the split datasets by tuning the parameter in the given grid. In what follows, we will show the detailed experiments individually.

\begin{figure}
\centering
\begin{tabular}{cc}
\includegraphics[width=0.43\textwidth]{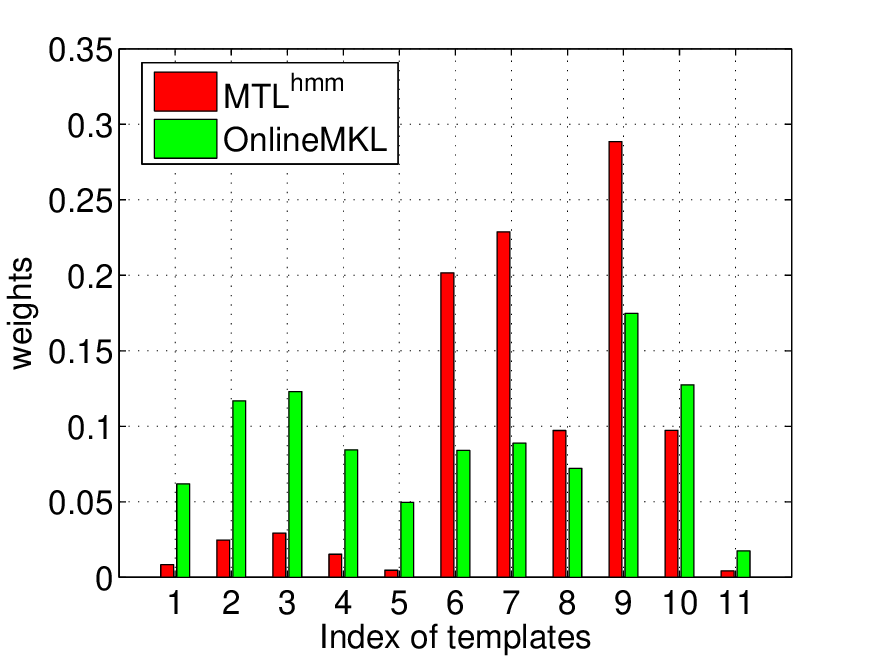} &
\includegraphics[width=0.43\textwidth]{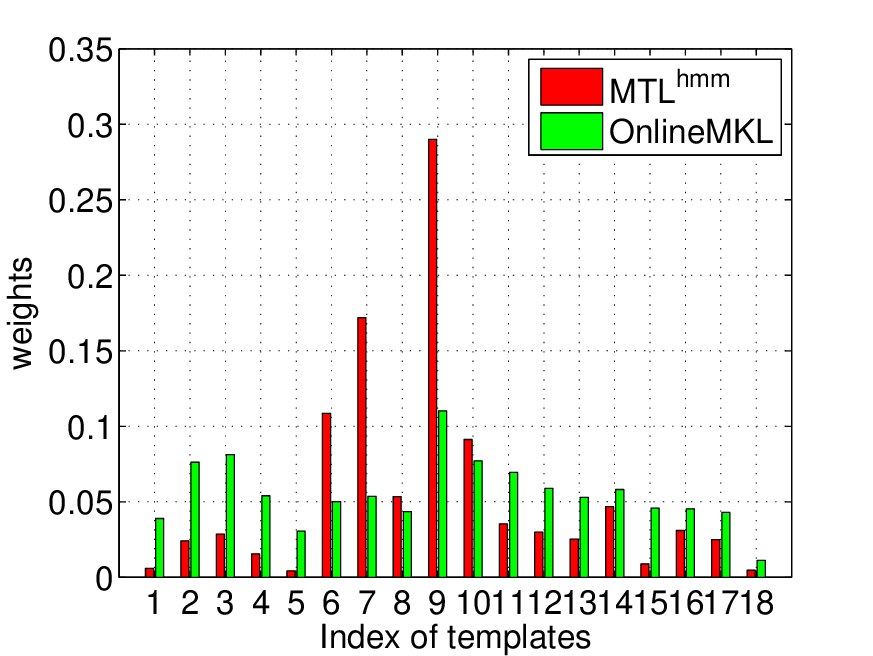}\\
(a) TP1 &(b)TP2\\
\end{tabular}
 \caption{ The learned template weights for Chinese word segmentation in TP1 and TP2 on PKU. }\label{fig:as_weight}
\end{figure}

\subsection{Chinese Word Segmentation}\label{sec:cws}

\begin{figure}
\begin{tabular}{cc}
\!\!\!\!\!\!\includegraphics[width=0.5\textwidth]{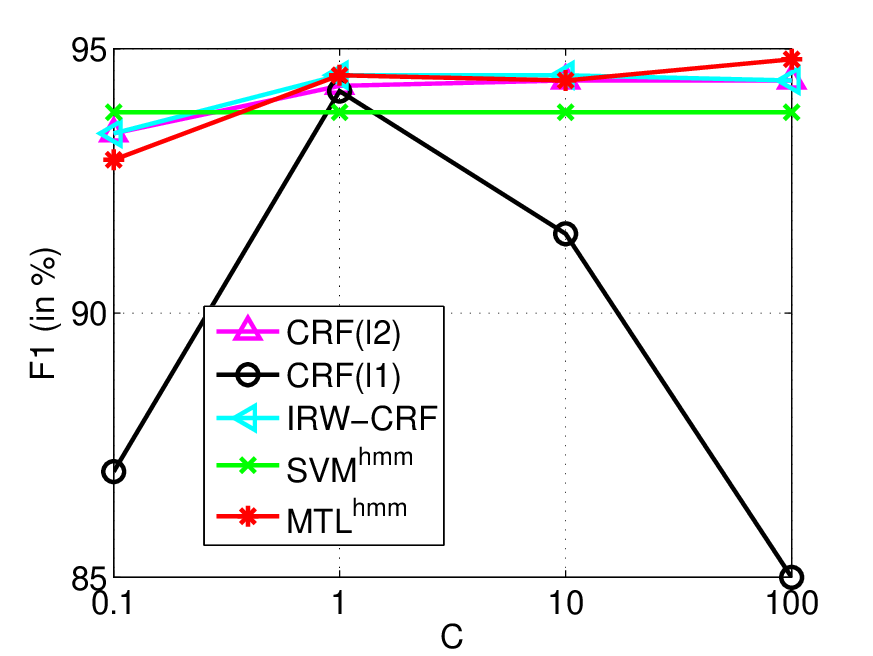} & \!\!\!\!\!\!\includegraphics[width=0.5\textwidth]{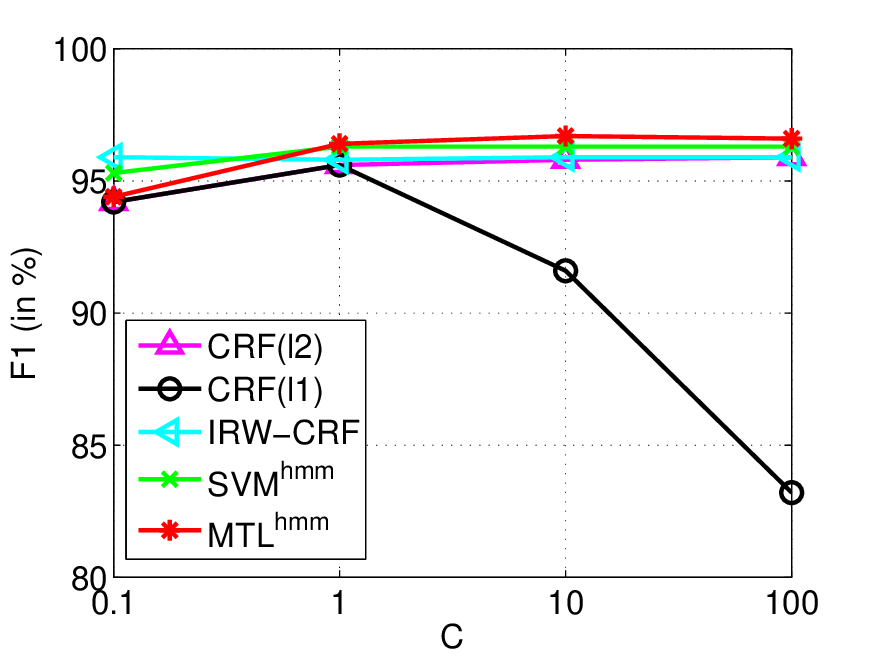} \\
\!\!\!\!\!\!(a) AS & \!\!\!\!\!\!(b) MSR\\
\!\!\!\!\!\!\includegraphics[width=0.5\textwidth]{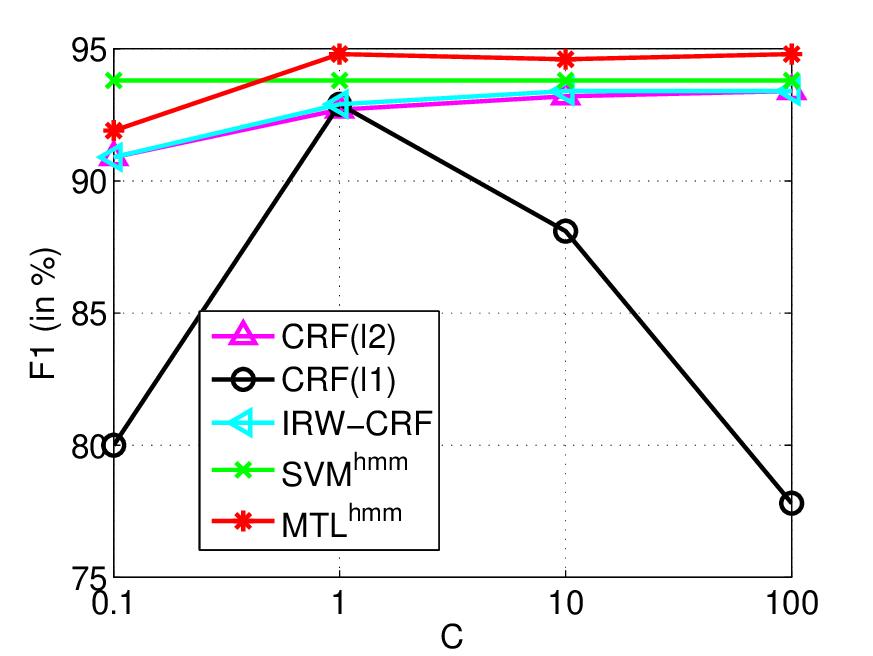} & \!\!\!\!\!\!\includegraphics[width=0.5\textwidth]{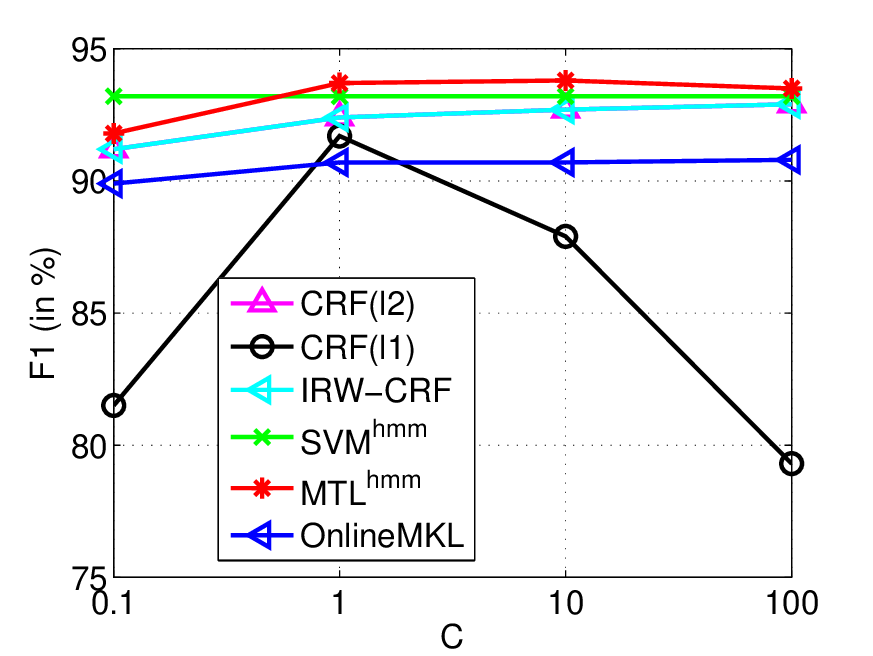} \\
\!\!\!\!\!\!(c) CityU & \!\!\!\!\!\!(d) PKU\\
\end{tabular}
 \caption{ F1 measure by varying $C$ in TP2 on the Chinese word segmentation datasets. }\label{fig:cws-different-c}
 \end{figure}

Chinese word segmentation has been an active area of research in computational linguistics for several decades. The second International Chinese Word Segmentation Bakeoff task\footnote{http://www.sighan.org/bakeoff2005/} provides the platform to evaluate different methods. It contains four corpora: Academia Sinica (AS), City University of Hong
Kong (CityU), Peking University (PKU) and Microsoft Research (MSR). The detailed corpus information is referred to \cite{Emerson05}. Table \ref{tab:seg_data_desciption} shows the benchmark splits of data and their associated features in terms of templates in Table \ref{tab:seg_templates} following the definition of templates in CRF++. Results are reported by the following measures: test recall (Recall), test precision (Prec), balanced F score (F1), and recall on in-vocabulary words ($\textrm{R}_{iv}$). We use three label notations such as B, I, E (B is the beginning character of a word; I is the inner character of a word; E is the end character of a word) to transform Chinese word segmentation task into a label sequence learning problem. To examine whether or not improper or redundant templates could affect sequence prediction performance, we divided the set of templates into two categories: TP1 and TP2. TP1 is frequently used in the literature \cite{Peng04}. In TP2, we deliberately enumerate all the possible bi-gram templates in the window size [-2,+2]. This construction of templates has a high probability to include redundant information.

\subsubsection{Model Stability and Training Time}

We empirically study the sensitivity of performances and the time complexities of each algorithms on the two smallest datasets PKU and CityU. By merging the training and testing data in Table \ref{tab:seg_data_desciption}, we randomly generate  $80\%$ training and $20\%$ testing splits five times. Results on PKU and CityU in Table \ref{tab:five-random-results} show that the standard deviations are less than $0.2\%$. The average CPU times on PKU dataset are shown in Figure \ref{fig:time-pku}. We can observe that our proposed method $\textrm{MTL}^{hmm}$ is slower than CRFs and $\textrm{SVM}^{hmm}$ in the range of large $C$ values, but it is more than $10$ times faster than OnlineMKL, and comparable to IRW-CRF. This is consistent with the complexity analysis in Section \ref{sec:complexity}. Moreover, the experiments of OnlineMKL cannot be finished in the reasonable time for the other three large scale datasets.

\subsubsection{Performance Analysis}

The testing results on the standard split of the datasets are reported in Table \ref{tab:seg}. In TP1, except OnlineMKL, the rest four algorithms have similar F1 score with
a difference at most $0.5\%$, but CRF($\ell_2$) is $1\%$ higher than $\textrm{SVM}^{hmm}$ on AS. IRW-CRF slightly outperforms CRF($\ell_2$) at most cases. However, $\textrm{MTL}^{hmm}$ always demonstrates the highest $R_{iv}$, which means weighting strategy
is more useful to predict the known words correctly.  Although $\textrm{MTL}^{hmm}$ and $\textrm{SVM}^{hmm}$ use the same set of features,
$\textrm{MTL}^{hmm}$ is better than $\textrm{SVM}^{hmm}$ according to F1 score.
This shows that the proposed weighting template strategy is helpful to boost the performance. However, OnlineMKL performs the worst among all algorithms.
It may require some additional  online-to-batch strategies~\cite{Dekel05,Dekel2008} for OnlineMKL to achieve the optimal batch-mode structured prediction performance.
In TP2, the similar phenomena could be observed, but the difference of F1 score among different methods are enlarged. For example, for CityU dataset, $\textrm{MTL}^{hmm}$ is $2.1\%$ higher than CRFs.
$\textrm{MTL}^{hmm}$ consistently shows the best F1 score.

By comparing the results between TP1 and TP2, we find that adding more templates, such as templates $U10$-$U16$,
degrades the performance of CRFs and $\textrm{SVM}^{hmm}$, but $\textrm{MTL}^{hmm}$ and OnlineMKL obtain slight improvements. Therefore, the performance of weighting strategy in $\textrm{MTL}^{hmm}$ and OnlineMKL is not greatly affected by the arbitrarily added templates. However, CRF($\ell_2$) and $\textrm{SVM}^{hmm}$ do not consider this information, and CRF($\ell_1$) works poorly on CityU and PKU. Moreover, OnlineMKL demonstrates  worse performance than $\textrm{MTL}^{hmm}$ more than $3\%$ with respect to F1 on PKU. This is due to the fact that $\textrm{MTL}^{hmm}$ can effectively identify the important templates from  a given set of templates, and can remove unimportant templates simultaneously; while OnlineMKL may not be so effective in this case.

\subsubsection{Weighting on Templates}

The above statements can be justified by the learned template weights. Figure \ref{fig:as_weight} shows an example of the learned weights of templates on PKU in the setting of TP1 and TP2. We can observe that the weights learned by $\textrm{MTL}^{hmm}$ has a higher standard deviation than that of OnlineMKL. The small standard deviation means the weights are more likely to be average weights. Hence, $\textrm{MTL}^{hmm}$ prefers sparser weights than OnlineMKL. Even using more epochs, OnlineMKL still cannot improve the sparsity of weights and prediction performance but takes more training time. Together with the better performance, this implies that $\textrm{MTL}^{hmm}$ trained in batch mode is more effective than OnlineMKL in online mode in this task.

\subsubsection{Sensitivity of Parameter $C$}

Figure \ref{fig:cws-different-c} shows the variations of testing F1 measure over the range of $C$ used in the experimental setting \footnote{Due to expensive training costs on large structured prediction tasks, we cannot afford a finer grid of $C$ parameter in our experiments.}. $\textrm{MTL}^{hmm}$ obtain better results than the rest of algorithms in the range with the large $C$, i.e. $C \geq 1$. Therefore, weighting groups of features can boost the F1 measure on all four Chinese word segmentation datasets in the framework of MTL$^{struct}$, but OnlineMKL degrades the performance due to the online mode. CRF($\ell_1$) varies greatly in terms of different $C$'s, which implies that aggressively removing features could degrade prediction performance \footnote{ A finer grid of $C$ parameter for tuning may improve the prediction performance of CRF($\ell_1$) on these four datasets.}.

\begin{table}
\caption{The overall performance of the comparing methods on Spanish and Dutch for named entity recognition. The best results across different methods are shown in bold.} \label{tab:ner}
\begin{center}
\begin{tabular}{l|ccc|ccc}
\hline
\multirow{2}{*}{Method}& \multicolumn{3}{|c|}{Validation set} & \multicolumn{3}{|c}{Test set}\\ \cline{2-7}
        & Prec   & Recall     & F1        & Prec   & Recall     & F1 \\
        \hline
       \hline
 CRF ($\ell_2$)       &74.38	&55.97	&63.88	&78.16	&60.72	&68.34 \\
 CRF ($\ell_1$)       &73.32	&58.34	&64.98	&77.41	&64.23	&70.21 \\
 IRW-CRF              &74.08    &56.23  &63.93  &77.17  &61.03  &68.39 \\
 $\textrm{SVM}^{hmm}$ &72.23	&57.01	&63.72	&74.83	&61.23	&67.35 \\
 OnlineMKL            &72.64	&62.64	&67.27	&75.06	&65.52	&69.97 \\
 $\textrm{MTL}^{hmm}$ &\textbf{77.47}	&\textbf{63.21}	&\textbf{69.62}	&\textbf{79.26}	&\textbf{68.39}	 &\textbf{73.42} \\
\hline
\multicolumn{7}{c}{(a) Spanish (ESP)}\\
       \hline
 CRF ($\ell_2$)       &71.56	&40.79	&51.96	&74.53	&46.41	&57.20 \\
 CRF ($\ell_1$)       &72.64	&43.85	&54.68	&74.93	&51.10	&60.76 \\
 IRW-CRF              &71.62    &41.09  &52.22  &73.68  &46.54  &57.05 \\
 $\textrm{SVM}^{hmm}$ &58.57	&39.98	&47.52	&60.15	&44.15	&50.92 \\
 OnlineMKL            &63.69	&43.39	&51.61	&68.09	&49.43	&57.28 \\
 $\textrm{MTL}^{hmm}$ &\textbf{82.72}	&\textbf{48.32}	&\textbf{61.00}	&\textbf{85.63}	&\textbf{56.08}	 &\textbf{67.77}\\
\hline
\multicolumn{7}{c}{(b) Dutch (NED)}\\
\hline
\end{tabular}
\end{center}
\end{table}

\begin{figure}
\centering
\begin{tabular}{cc}
\includegraphics[width=0.5\textwidth]{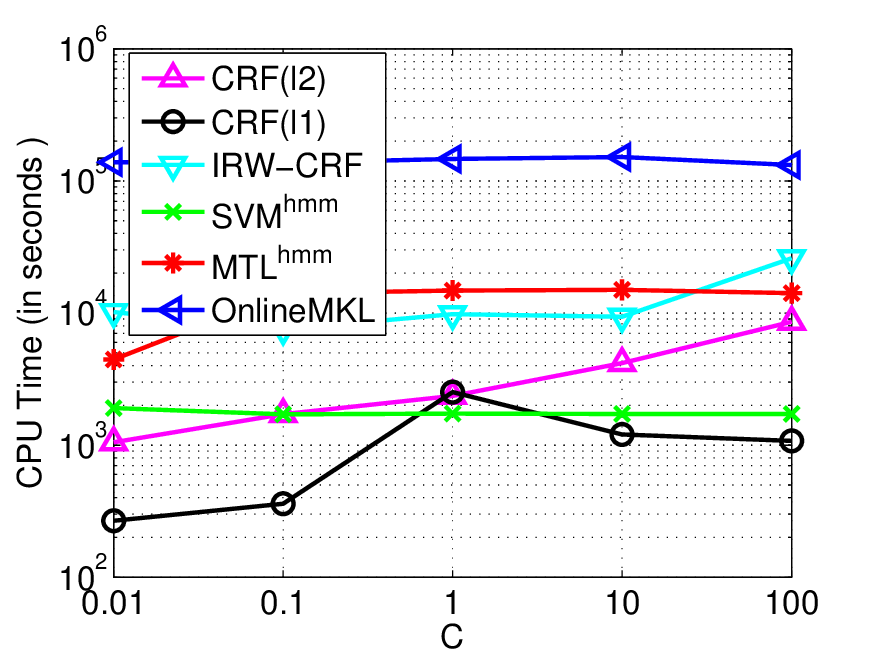}\!\!\!&\!\!\!\includegraphics[width=0.5\textwidth]{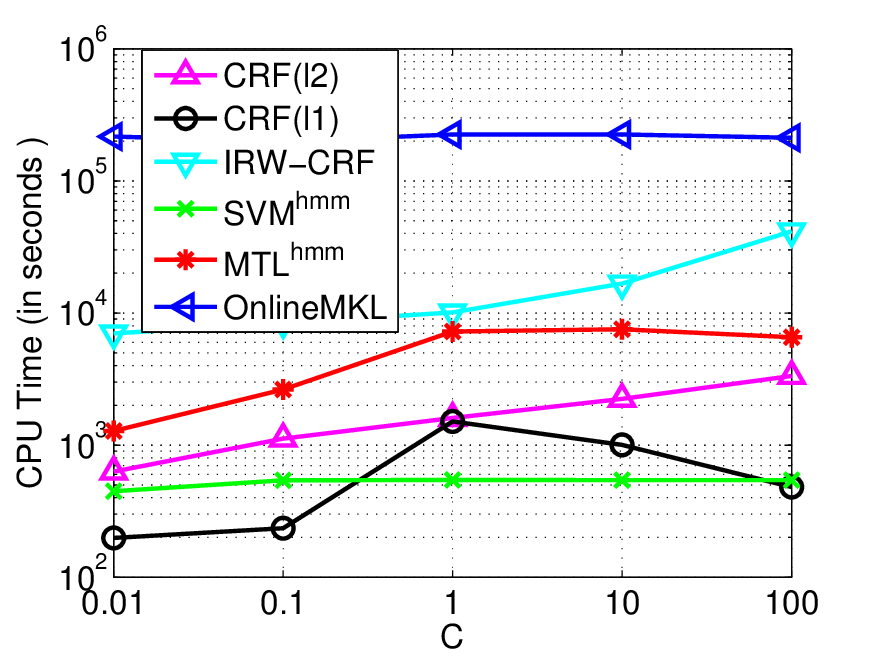}\\
(a) ESP & (b) NED \\
\end{tabular}
\caption{The CPU time on Spanish and Dutch dataset by varying $C$.}
\label{fig:time-ner}
\end{figure}

\begin{figure}
\centering
\begin{tabular}{cc}
\!\!\!\!\!\!\!\!\!\!\!\!\includegraphics[width=0.45\textwidth,height=1.9in]{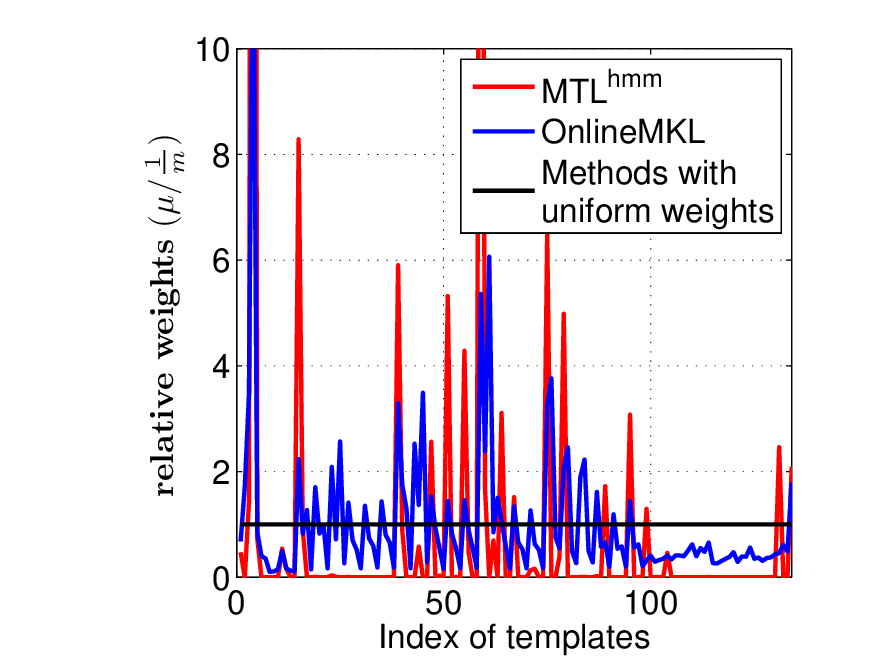} &

\!\!\!\!\!\!\!\!\!\!\!\!\includegraphics[width=0.45\textwidth,height=1.9in]{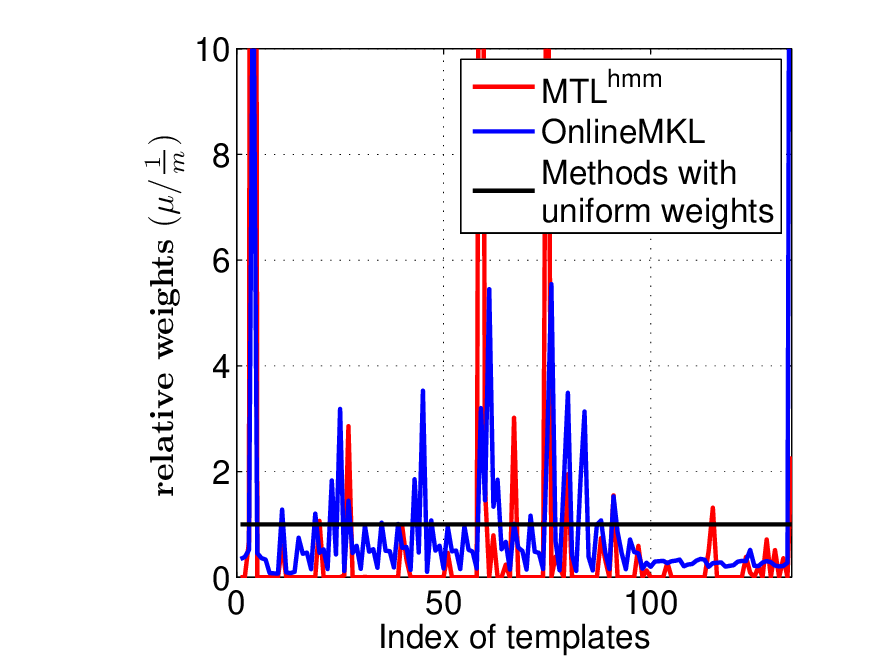}\\
(a) ESP &(b) NED
\end{tabular}
\caption{ The relative learned template weights for Named entity recognition on Spanish and Dutch. The horizontal line with $1$ stands for methods with the average weights. }\label{fig:ner_weight}
\end{figure}

\subsection{Named Entity Recognition} \label{sec:ner}

We evaluate all methods on the CoNLL-2002 language-independent named entity recognition shared task\footnote{http://www.cnts.ua.ac.be/conll2002/ner/}. The data consists of three files per language: training, validation (testa), and testing (testb). Spanish (ESP) $(8,323 / 1,915 / 1,517)$ and Dutch (NED) $(15,806 / 2,895 / 5,159)$  with the number of sentences in each file are used in this experiment. Similar to TP2 in Table \ref{tab:seg_templates}, we enumerate in the range of $[-3,3]$ the combination of uni-gram and bi-gram templates in terms of words and part-of-speech (POS) respectively, the combination of tri-gram on POS only, and a state-state transition template. There are $134$ templates in total, and the number of features are $5,926,876$ and $4,710,332$ for ESP and NED, respectively.

\subsubsection{Performance Analysis on All Templates}

Table \ref{tab:ner} shows the overall performance of different methods on Spanish and Dutch data, respectively. We observe that CRF ($\ell_2$) and  $\textrm{SVM}^{hmm}$ perform worse than other methods. This implies that template (feature) selection improves performance by pruning the noisy or redundancy templates (features). Another observation is that $\textrm{MTL}^{hmm}$ outperforms CRF ($\ell_1$), IRW-CRF  and OnlineMKL with more than $3\%$ on ESP and $7\%$ on NED in terms of F1 on the testing data. Figure \ref{fig:time-ner} shows the training times of each method where $\textrm{MTL}^{hmm}$ is slower than CRFs and  $\textrm{SVM}^{hmm}$, but it is more than $10$ times faster than  OnlineMKL. Figure \ref{fig:ner_weight} shows the weights learned by OnlineMKL and $\textrm{MTL}^{hmm}$ on both datasets. If we consider the template with its weight less than a relative small value such as $10^{-5}$ is redundant, $\textrm{MTL}^{hmm}$ can remove $75$ and $97$ templates for ESP and NED, respectively, but OnlineMKL cannot remove any template. On both datasets, $\textrm{MTL}^{hmm}$ obtains much sparser weights than OnlineMKL, and also selects a quite different subset of templates for different language; while OnlineMKL selects similar subset of templates due to the similar curves. All the observations imply that $\textrm{MTL}^{hmm}$ is more promising than other methods.

\subsubsection{Performance on the Selected Templates by MTL$^{hmm}$}

\begin{table}[h]
\caption{The overall performance of the comparing methods on Spanish and Dutch for named entity recognition with the selected subset of templates.  The best results across different methods are shown in bold.}
\label{tab:ner2}
\begin{center}
\begin{tabular}{l|ccc|ccc}
\hline
\multirow{2}{*}{Method}& \multicolumn{3}{|c|}{Validation set} & \multicolumn{3}{|c}{Test set}\\ \cline{2-7}
        & Prec   & Recall     & F1        & Prec   & Recall     & F1 \\
        \hline
       \hline
 CRF ($\ell_2$)       &\textbf{78.00}   &61.03  &68.48  &\textbf{81.76}  &67.52  &73.96 \\
 CRF ($\ell_1$)       &76.58   &60.34  &67.50  &80.75  &66.70  &73.00 \\
 IRW-CRF              &76.36   &\textbf{63.90}  &\textbf{69.58}  &81.04  &\textbf{70.50}  &\textbf{75.40} \\
 $\textrm{SVM}^{hmm}$ &74.39   &62.82  &68.12  &78.17  &68.50  &73.02 \\
 OnlineMKL            &73.17   &63.60  &68.05  &75.58  &68.19  &71.70 \\
 $\textrm{MTL}^{hmm}$ &76.09   &62.32  &68.52  &79.96  &69.74  &74.50 \\
\hline
\multicolumn{7}{c}{(a) Spanish (ESP)}\\
       \hline
 CRF ($\ell_2$)       &77.66   &43.31  &55.61  &80.79  &50.57  &62.20 \\
 CRF ($\ell_1$)       &77.53   &43.92  &56.08  &80.05  &51.31  &62.53 \\
 IRW-CRF              & 79.22 &48.09  &59.85   &82.19   &54.78  &65.74       \\
 $\textrm{SVM}^{hmm}$ &70.16   &43.96  &54.05  &74.77  &50.90  &60.57 \\
 OnlineMKL            &67.63   &45.83  &54.64  &69.60  &51.99  &59.52 \\
 $\textrm{MTL}^{hmm}$ &\textbf{80.64}   &\textbf{47.94}  &\textbf{60.13}  &\textbf{85.06}  &\textbf{55.34}  &\textbf{67.06}\\
\hline
\multicolumn{7}{c}{(b) Dutch (NED)}\\
\hline
\end{tabular}
\end{center}
\end{table}

To further justify the effectiveness, we select a subset of $20$ templates according to the learned weights of MTL$^{hmm}$ for all methods. The overall performances are shown in Table \ref{tab:ner2}. We observe that the overall performances on the small subset of templates greatly outperform that on the full set for both CRF ($\ell_2$) and $\textrm{SVM}^{hmm}$ by more than $5\%$ over both datasets. CRF ($\ell_1$) and OnlineMKL also have around $2$-$3\%$ improvement.  $\textrm{MTL}^{hmm}$ on ESP has $1.02\%$ improvement, while a reduction of $0.71\%$ happens on NED. IRW-CRF in this case show the best performance on ESP, but worse on NED comparing with $\textrm{MTL}^{hmm}$. However, IRW-CRF performs worse on the full set of templates shown in Table \ref{tab:ner} which implies that IRW-CRF cannot tackle well the noisy templates. All these observations imply that $\textrm{MTL}^{hmm}$ is effective for  selecting templates and has a stable even improved performance in the case of noisy or non-informative templates.

\subsection{Dependency Parsing}

We perform comparisons on CoNLL-X shared task\footnote{http://nextens.uvt.nl/\~{}conll/post\_task\_data.html}: Multi-lingual Dependency Parsing, on Danish and Swedish where the number of sentences in the given files (training / testing) are $(5,125/313)$ and $(11,092/389)$, respectively.
We use the templates in MSTParser including: (1) POS (including CPOSTAG and POSTAG) trigrams: the POS of the head, that of the modifier and that of a word in between, for all distinct POS tags for the words between the head and the modifier. Each relative position from the head to the modifier can be considered as a different type of template. (2) The form of POS 4-gram: the POS of the head, modifier, word before/after head and word before/after modifier. (3) Two items: each template consists of two observations, e.g. head word, head POS/LEMMA, child word, and child POS/LEMMA. All templates are conjoined with the direction of attachment as well as the distance between the two words creating the dependency. For the distance between two words creating the dependency is longer than $10$, it is set to $10$; If between $10$ and $5$, it is $5$, otherwise it does not change.
There are 568 and 612 templates which generate $2,137,252$ and $2,292,216$ features for Danish and Swedish, respectively. Three measures are employed to evaluate the performance of comparing methods: complete sentence accuracy (ACC), unlabeled attachment score including punctuation (UASP), and unlabeled attachment score excluding punctuation (UAS).

\begin{table}
\caption{The results (MSTParser / MTL$^{parse}$) of the dependency parsing on Danish and Swedish. The best results across the compared methods are shown in bold. } \label{tab:parsing-tree-results}
\begin{center}
\begin{tabular}{l|c|c|c}
\hline
              &   ACC &  UASP & UAS \\
\hline Danish & 31.06 / \textbf{32.61}  &  86.93 / \textbf{87.08}    &   \textbf{88.80} / \textbf{88.80}  \\
\hline Swedish& 37.28 / \textbf{37.53 } &83.65 /\textbf{ 84.14}  &85.76 / \textbf{86.40}      \\
\hline
\end{tabular}
\end{center}
\end{table}

\begin{figure}
\begin{tabular}{cc}
\!\!\!\!\!\!\!\!\includegraphics[width=0.5\textwidth,height=1.4in]{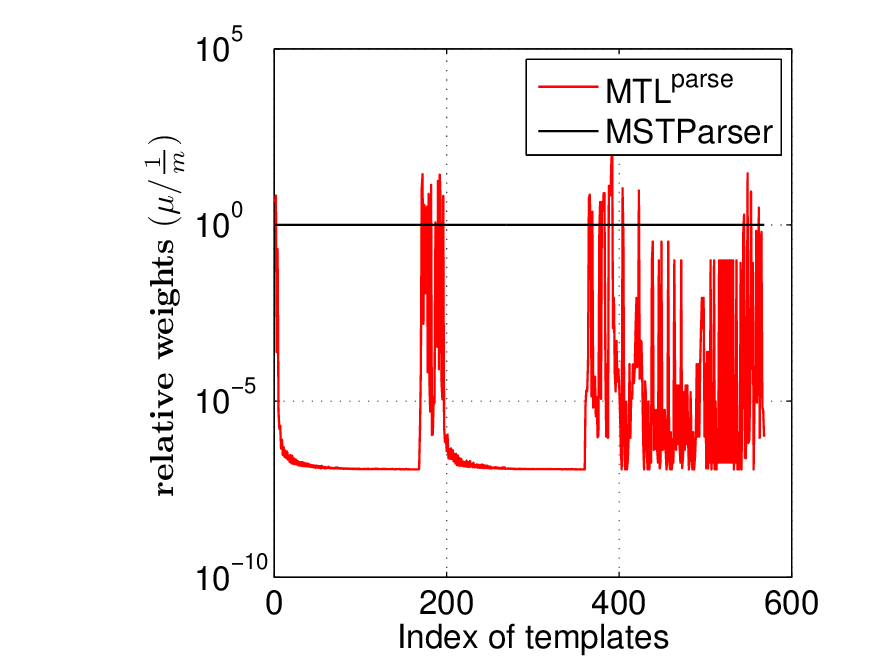}\!\!\!\!\!\!\!& \!\!\!\!\!\!\!\includegraphics[width=0.5\textwidth,height=1.4in]{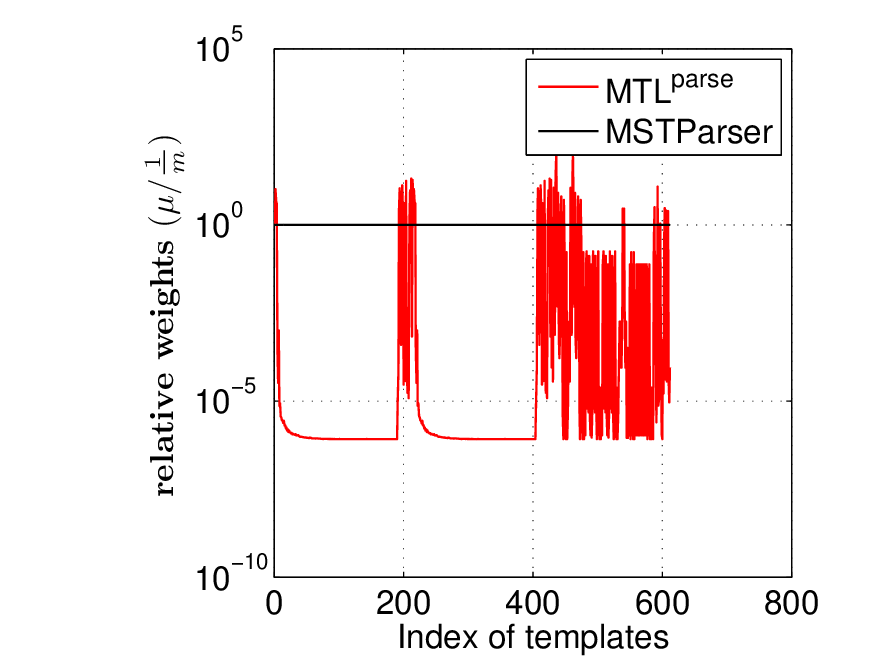}\\
(a) Danish & (b) Swedish
\end{tabular}
 \caption{ The relative learned template weights for Dependency Parsing on Danish and Swedish. The horizontal line with $10^0$ stands for methods with the average weights.}\label{fig:parsing_weight}
\end{figure}

The experimental results of comparisons over Danish and Swedish are shown in Table \ref{tab:parsing-tree-results}, and the corresponding learned weights of templates are presented in Figure \ref{fig:parsing_weight}. If the same selection strategy is used in named entity recognition task, MTL$^{parse}$ can remove $500$ and $508$ templates for Danish and Swedish, respectively. It implies that the huge redundancy exists in the given set of templates. The similar phenomena can be observed: MTL$^{parse}$ obtains comparable or even better results than MSTParser, and also achieves very sparse weights of templates. The weights of patterns in some senses could be helpful for the interpretation of semantic meaning of each language \cite{Martins10}.

\subsection{MTL$^{struct}$ with $p$-Block-Norm Regularization}

\begin{table}[t]
\caption{The overall results of $p$MTL$^{hmm}$ on PKU and ESP by varying $p$. The best results across the different $p$ are shown in bold.} \label{tab:p-block-MTL}
\begin{center}
\begin{tabular}{c|cccc|ccc}
\hline
\multirow{2}{*}{$p$}& \multicolumn{4}{|c|}{PKU} & \multicolumn{3}{|c}{ESP (Test set)}\\ \cline{2-8}
        & Recall   & Prec     & F1  & R$_{iv}$       & Prec   & Recall     & F1 \\
        \hline
        \hline
1   & \textbf{93.8} & 93.3 & \textbf{93.6} & \textbf{96.1} & 79.96 & 69.74 & 74.50\\
4/3 & 92.9 & \textbf{94.1} & 93.5 & 94.7 & \textbf{80.22} & \textbf{70.08} & \textbf{74.81}\\
2   & 92.6 & 93.4 & 93.0 & 94.9 & 79.08 & 67.35 & 72.75\\
4   & 90.8 & 91.3 & 91.1 & 93.4 & 74.59 & 59.54 & 66.22 \\
100 & 87.4 & 89.7 & 88.5 & 90.0 & 71.92 & 54.34 & 61.91\\
\hline
\multicolumn{8}{c}{(a) small set of templates}\\
\hline
1   & \textbf{94.0} & 93.6 & \textbf{93.8} & \textbf{96.2} & \textbf{79.26} & \textbf{68.39} & \textbf{73.42} \\
4/3 & 93.5 & \textbf{94.0} & \textbf{93.8} & 95.3 & 78.29 & 64.65 & 70.82 \\
2   & 92.6 & 93.7 & 93.1 & 94.3 & 75.52 & 61.11 & 67.56 \\
4   & 90.0 & 92.5 & 91.3 & 92.2 & 72.74 & 53.93 & 61.58 \\
100 & 89.7 & 91.7 & 90.7 & 91.9 & 65.08 & 54.51 & 59.33\\
\hline
\multicolumn{8}{c}{(b) large set of templates}\\
\hline
\end{tabular}
\end{center}
\end{table}

\begin{figure}[t]
\centering
\begin{tabular}{c}
\includegraphics[width=0.5\textwidth]{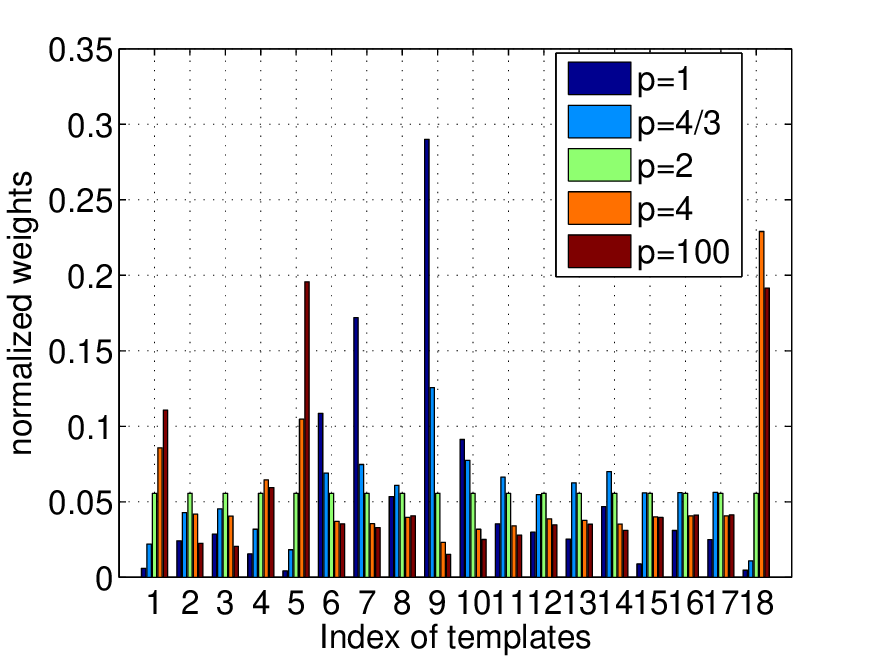} \\
(a) PKU \\
\includegraphics[width=0.5\textwidth]{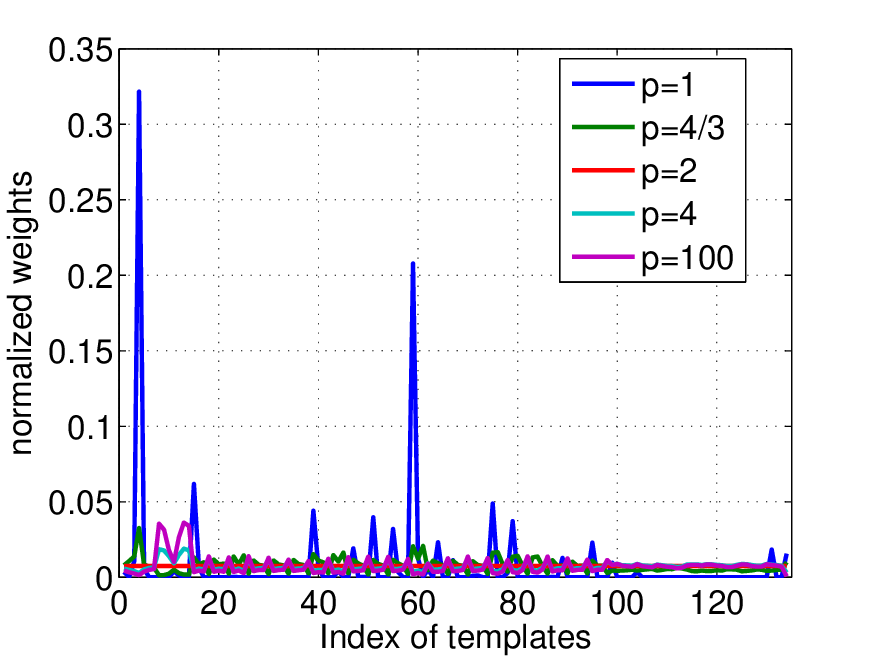}\\
 (b) ESP
\end{tabular}
 \caption{ The normalized weights of $p$MTL$^{hmm}$ on the large set of templates by varying $p$ values on PKU and ESP datasets, respectively. }\label{fig:phmmkl_weight}
\end{figure}

Experiments of $p$\textbf{MTL}$^{hmm}$ are carried out on PKU and ESP (Test set), and their corresponding templates are employed (see Section \ref{sec:cws} and Section \ref{sec:ner}, respectively). The empirical results are shown in Table \ref{tab:p-block-MTL}, where $p=1$ is MTL$^{hmm}$ trained by Algorithm \ref{algo:smkl}, and $p$=$100$ is used to approximate infinity-block norm. We observe that F1 score decreases when increasing $p$, except that $p$=$3/4$ obtains the best performance on ESP (Test set). For large $p$, the performance degrades greatly. The corresponding weights on the large set of templates are shown in Figure \ref{fig:phmmkl_weight} where the weights are normalized by $l_1$ norm due to the scaling problem. From these observations, $1 \leq p < 2$ is recommended for sequence learning tasks studied in this paper.

\section{Conclusion} \label{sec:conclusion}
Structured prediction is an important modeling strategy for various real world applications. Instead of modeling the structure of specific application, in this paper, we explore the underlying feature structure over the input-output pairs which contributes to the success of discriminative models, such as CRFs and Structural SVMs. Structured prediction algorithms take more effort on how to model the interdependence among the output variables, but less consideration is taken on the feature engineering which is a non-trivial and tedious task for general users. To alleviate this issue, we propose a Multiple Template Learning (MTL$^{struct}$) paradigm to learn both the weight of each template and the structured prediction model in the batch mode, simultaneously. Learning the weights of these groups is formulated as a Multiple Kernel Learning (MKL) problem. We proposed to solve this MKL problem in the primal by an efficient cutting plane algorithm, and its convergence is presented. We further extend it to the  $p$-block norm regularization and the modified algorithm is readily obtained.
Two special cases are explored in our proposed MTL$^{struct}$ framework, i.e. sequence labeling and dependency parsing. Extensive experimental results demonstrate that learning structured prediction model with weighting template can automatically interpret the importance of each templates, so users can arbitrarily define numerous templates without cautions.
Both theoretical analysis and empirical results verified that our MTL$^{struct}$ is more efficient and performs much better than OnlineMKL.
In future, we will extend the proposed framework for structured prediction with regression outputs.

\appendices
\section{The Conic Dual of Problem (\ref{op:margin-scaling-mkl-one-slack-subproblem}) } \label{proof:qcqp}

\begin{proof}
Without loss of generality, we denote $\textbf{w}=[\textbf{w}_1;\ldots;\textbf{w}_m ]$ and $\textbf{p}^r = [\textbf{p}^r_1;\ldots;\textbf{p}^r_m]$. Problem (\ref{op:margin-scaling-mkl-one-slack-subproblem}) becomes
\begin{eqnarray*}
\min_{\textbf{w}, \xi \geq 0} && \frac{1}{2} \Bigg( \sum_{j=1}^m \|\textbf{w}_j\| \Bigg)^2 + C \xi \\
\textrm{s.t.} && \xi \geq q^r + \textbf{w}^T \textbf{p}^r, \forall r=1,\ldots,s
\end{eqnarray*}
By introducing a new variable $u \in \mathbb{R}$ and moving out summation operator from objective to be a constraint, we can obtain the equivalent optimization problem as
\begin{eqnarray*}
\min_{\textbf{w}, \xi \geq 0} && \frac{1}{2} u^2 + C \xi \\
\textrm{s.t.} && \xi \geq q^r + \textbf{w}^T \textbf{p}^r, \forall r=1,\ldots,s \\
              && \sum_{j=1}^m \|\textbf{w}_j\| \leq u.
\end{eqnarray*}
We can further simplify the above problem by introducing another variables $\rho \in \mathbb{R}^m$ such that $\|\textbf{w}_j\| \leq \rho_j$, $\forall j=1,\ldots,m+1$ to be
\begin{eqnarray*}
\min_{\textbf{w}, \xi \geq 0} && \frac{1}{2} u^2 + C \xi \\
\textrm{s.t.} && \xi \geq q^r + \textbf{w}^T \textbf{p}^r, \forall r=1,\ldots,s \\
              && \sum_{j=1}^m \rho_j \leq u, ||\textbf{w}_j|| \leq \rho_j, \forall j=1,\ldots,m.
\end{eqnarray*}
We know that $\|\textbf{w}_j\| \leq \rho_j$ is a second-order cone constraint. Following the recipe of \cite{Boyd04}, the self-dual cone $\|\textbf{v}_j\|_2 \leq \eta_j, \forall j=1,\ldots,m$ can be introduced to form the Lagrangian function:
$\mathcal{L}(\textbf{w},\xi,u,\rho; \alpha,\tau,\gamma,\textbf{v},\eta) = \frac{1}{2} u^2 + C \xi - \sum_{r=1}^s \alpha_r (\xi - q^r - \textbf{w}^T \textbf{p}^r )
- \tau \xi + \gamma \big( \sum_{j=1}^m \rho_j - u \big) - \sum_{j=1}^m (\langle \textbf{v}_j, \textbf{w}_j \rangle + \eta_j \rho_j)$,
with dual variables $\alpha_r \in \mathbb{R}_+$, $\tau \in \mathbb{R}_+$, $\gamma \in \mathbb{R}_+$. The derivatives of the Lagrangian with respect to the primal variables have to vanish which leads to the following KKT conditions:
$\textbf{v}_j = \sum_{r=1}^s \alpha_r \textbf{p}^r_j, \forall j=1,\ldots,m $,
$C - \sum_{r=1}^s \alpha_r - \tau = 0$,
$u = \gamma$, and
$\gamma = \eta_j, \forall j=1,\ldots,m$.
By substituting all the primal variables with dual variables by above KKT conditions, we can obtain the following dual problem,
\begin{eqnarray*}
\max_{\alpha,\gamma} && -\frac{1}{2} \gamma^2 + \sum_{r=1}^s \alpha_r q^r \\
\textrm{s.t.} && \Big\| \sum_{r=1}^s \alpha_r \textbf{p}^r_j \Big\| \leq \gamma, \forall j=1,\ldots,m\\
              &&\sum_{r=1}^s \alpha_r \leq C, \alpha_r \geq 0, \forall r=1,\ldots,m
\end{eqnarray*}
By setting $\theta = \frac{1}{2}\gamma^2$ and $\mathcal{A}_s=\{\sum_{r=1}^s \alpha_r \leq C, \alpha_r \geq 0, \forall r=1,\ldots,s\}$, we can reformulate the above problem as
\begin{eqnarray*}
\max_{\theta, \alpha \in \mathcal{A}_s} -\theta + \sum_{r=1}^s \alpha_r q^r : \frac{1}{2} \alpha^T Q^j \alpha \leq \theta, \forall j=1,\ldots,m
\end{eqnarray*}
where $Q_{r,r'}^j = \langle \textbf{p}_j^r, \textbf{p}_j^{r'} \rangle$. According to the property of self-dual cone, we can obtain the primal solution from its dual as $\textbf{w}_j = -\mu_j \textbf{v}_j = -\mu_j \sum_{r=1}^s \alpha_r \textbf{p}^r_j$ where $\mu_j$ is the dual variable of the $j^{th}$ quadratic constraint such that $\sum_{j=1}^m \mu_j = 1,\mu_j \in \mathbb{R}_+, \forall j=1,\ldots,m$~\cite{Bach04}.
\end{proof}

\section{Proof of the Theorem 2} \label{app:proof-theorem2}

\begin{proof}
Since there are $m$ quadratic constraints, the dual objective of MTL$^{struct}$ can be reformulated as
\begin{eqnarray*}
 \max_{\alpha \in \mathcal{A}_s} \min_{ j=1,\ldots,m} \Theta_{\textbf{d}}(\alpha) = -\frac{1}{2} \sum_{r=1}^s \sum_{r'=1}^s  \alpha_r  \alpha_{r'} Q^j_{r,r'} +  \sum_{r=1}^s \alpha_r q^r.
\end{eqnarray*}
We consider each group of features at one time:
\begin{displaymath}
\max_{\alpha \in \mathcal{A}_s} \Theta_j(\alpha),
\end{displaymath}
where $Q^j$ is positive semi-definite matrix, and derivative $\partial \Theta_j(\alpha) = q^j - Q^j \alpha$. The Lemma 2 in \cite{Joachims09} states that a line search starting at $\alpha$ along an ascent direction $\eta$ with maximum step-size $C > 0$ improves the objective by at least
$\max_{0 \leq \beta \leq C}  \big\{\Theta_j(\alpha + \beta \eta) - \Theta_j(\alpha) \big\}
\geq \frac{1}{2} \min \left\{ C, \frac{\partial \Theta_j(\alpha)^T \eta}{\eta^T Q^j \eta} \right\} \partial \Theta_j(\alpha)^T \eta.$
If we consider subgradient descent method, the line search along the subgradient of objective is $\partial \Theta_{j^*} (\alpha)$ where $j^* = \arg \min_{j \in \{1,\ldots,m\}} \Theta_j(\alpha)$. Therefore, the maximum improvement is
\begin{eqnarray}
&&\max_{0 \leq \beta \leq C}  \{\Theta_{j^*}(\alpha + \beta \eta) - \Theta_{j^*}(\alpha) \} \nonumber\\
&\geq& \frac{1}{2} \min \left\{ C, \frac{\partial \Theta_{j^*}(\alpha)^T \eta}{\eta^T Q^{j^*} \eta} \right\} \partial\Theta_{j^*}(\alpha)^T \eta \nonumber\\
&\geq& \frac{1}{2} \min_{j \in \{1,\ldots,m\}} \left\{ C, \frac{\partial \Theta_j(\alpha)^T \eta}{\eta^T Q^j \eta} \right\} \partial \Theta_j(\alpha)^T \eta. \label{eq:proof}
\end{eqnarray}
We can observe that it is a special case of \cite{Joachims09} if there is only one template. According to Theorem 5 in \cite{Joachims09}, for a newly added constraint $\widehat{\mathcal{Y}} \in \mathcal{W}$ and some $\gamma_j > 0$, we can obtain $\partial \Theta_j(\alpha)^T \eta = \gamma_j$ by setting the ascent direction $\eta_{\widehat{\mathcal{Y}}}=1$ for the newly added $\widehat{\mathcal{Y}}$ and $\eta_r = -\frac{1}{C} \alpha_r$ for the others.
 Here, we  set $\gamma = \min_{j \in \{1,\ldots,m\}} \gamma_j$ so as to be the lower bound of $\partial \Theta_j(\alpha)^T \eta, \forall j=1,\ldots,m$.
In addition, the upper bound for $\eta^T Q^j \eta \leq 4 R^2, \forall j=1,\ldots,m+1$ can also be obtained by the fact that
 $\eta^T Q^j \eta = Q^j_{\widehat{\mathcal{Y}},\widehat{\mathcal{Y}}} - \frac{2}{C} \sum_{r=1}^s \alpha_{r} Q^j_{r,\widehat{\mathcal{Y}}} + \frac{1}{C^2} \sum_{r=1}^s \sum_{r'=1}^s \alpha_r \alpha_{r'} Q^j_{r,r'} \leq R^2 + \frac{2}{C} CR^2 + \frac{1}{C^2} C^2 R^2 = 4 R^2, \forall j=1,\ldots,m$.
By substituting them back to (\ref{eq:proof}), the similar result shows the increase of the objective is at least
\begin{displaymath}
\min \left\{ \frac{C \gamma}{2}, \frac{\gamma^2}{8R^2} \right\}.
\end{displaymath}
Moreover, the initial optimality gap is at most $C \Delta$ when $\textbf{w}=\textbf{0}$. Following the remaining derivation in \cite{Joachims09}, the overall bound results are obtained.
\end{proof}

\section{Proof of Proposition \ref{prop:pmkl}} \label{proof:pmkl}

\begin{proof}
The Lagrangian dual problem of (\ref{op:margin-scaling-pmkl-one-slack-subproblem}) is formulated with dual variables $\alpha_r \geq 0, \forall r=1,\ldots,s$, and $\tau \geq 0$ as
\begin{eqnarray*}
\max_{\alpha \geq 0} \min_{\mathbf{w},\xi \geq 0} \frac{1}{2} ||\mathbf{w}||_{2,p}^2  + C \xi - \sum_{r=1}^s \alpha_r ( \xi - q^r - \sum_{j=1}^m \textbf{w}_j^T \textbf{p}_j^r ) - \tau \xi
\end{eqnarray*}
where $||\mathbf{w}||_{2,p}^2 = \big( \sum_{j=1}^m ||\textbf{w}_j||^p\big)^{2/p}$. Similar to the derivation of Appendix A, we can obtain the following KKT conditions
\begin{eqnarray}
\mathbf{w}_j = - ||\mathbf{w}||_{2,p}^{p-2} ||\mathbf{w}_j||^{2-p} \sum_{r=1}^s \alpha_r \textbf{p}_j^r, \forall j=1,\ldots, m  \label{append:kkt-w}\\
\sum_{r=1}^s \alpha_r \leq C, \alpha_r \geq 0, \forall r=1,\ldots,s. \nonumber
\end{eqnarray}
According to the Fenchel-Legendre conjugate function~\cite{Kloft2010}, we have the following equality
\begin{displaymath}
\max_{\mathbf{w}} -\frac{1}{2} \|\mathbf{w}\|_{2,p}^2 + \sum_{j=1}^m \textbf{w}_j^T \Big(-\sum_{r=1}^s \alpha_r \textbf{p}_j^r \Big) = \frac{1}{2}\Big\| -\sum_{r=1}^s \alpha_r \textbf{p}_j^r \Big\|_{2,p^*}^2
\end{displaymath}
such that $\frac{1}{p^*} + \frac{1}{p} = 1$. Substituting them back to the Lagrangian function, we reformulate the dual problem as
\begin{eqnarray*}
\max_{\alpha \in \mathcal{A}_s} -\frac{1}{2}\left|\left| \sum_{r=1}^s \alpha_r \textbf{p}_j^r \right|\right|_{2,p^*}^2 + \sum_{r=1}^s \alpha_r q^r
\end{eqnarray*}
which is equivalent to the Problem (\ref{op:pmkl-dual}).

According to (\ref{append:kkt-w}), we can obtain the following equation
\begin{small}
\begin{eqnarray*}
||\mathbf{w}_j|| = ||\mathbf{w}||_{2,p}^{\frac{p-2}{p-1}} \left|\left|\sum_{r=1}^s \alpha_r \textbf{p}_j^r\right|\right|^{\frac{1}{p-1}}, \forall j=1,\ldots, m.
\end{eqnarray*}
\end{small}\noindent
We further substitute it back to (\ref{append:kkt-w}), we obtain the primal solution
\begin{small}
\begin{eqnarray*}
\mathbf{w}_j = - ||\mathbf{w}||_{2,p}^{ \frac{p-2}{p-1}} \left|\left| \sum_{r=1}^s \alpha_r \textbf{p}_j^r \right|\right|^{ \frac{2-p}{p-1}} \sum_{r=1}^s \alpha_r \textbf{p}_j^r , \forall j=1,\ldots, m.
\end{eqnarray*}
\end{small}\noindent
Let $\mu_j = ||\mathbf{w}||_{2,p}^{ \frac{p-2}{p-1}} \left|\left| \sum_{r=1}^s \alpha_r \textbf{p}_j^r \right|\right|^{ \frac{2-p}{p-1}}, \forall j=1,\ldots, m$, so the primal solution can be simplified as
$\mathbf{w}_j = - \mu_j \sum_{r=1}^s \alpha_r \textbf{p}_j^r$.
According to the definition of the mixed norm, we can have the following derivations,
\begin{small}
\begin{eqnarray*}
||\mathbf{w}||_{2,p}^{ \frac{p-2}{p-1}}\!\!\! &=&\!\!\! \left( \sum_{j=1}^m ||\textbf{w}_j||^p\right)^{ \frac{p-2}{p(p-1)}} \\
                     \!\!\!&=&\!\!\! \left( \sum_{j=1}^m  \left(  ||\mathbf{w}||_{2,p}^{\frac{p-2}{p-1}} \left|\left|\sum_{r=1}^s \alpha_r \textbf{p}_j^r\right|\right|^{\frac{1}{p-1}} \right)^p \right)^{\frac{p-2}{p(p-1)}} \\
                     \!\!\!&=&\!\!\! \left(||\mathbf{w}||_{2,p}^{\frac{p-2}{p-1}} \right)^{\frac{p-2}{p-1}} \left(\sum_{j=1}^m \left|\left|\sum_{r=1}^s \alpha_r \textbf{p}_j^r\right|\right|^{\frac{p}{p-1}} \right)^{\frac{p-2}{p(p-1)}}.
\end{eqnarray*}
\end{small}\noindent
Hence, we can obtain the representation of the mixed norm with respect to $\alpha$ as
\begin{small}
\begin{eqnarray*}
||\mathbf{w}||_{2,p}^{ \frac{p-2}{p-1}} = \left(\sum_{j=1}^m \left|\left|\sum_{r=1}^s \alpha_r \textbf{p}_j^r\right|\right|^{\frac{p}{p-1}} \right)^{\frac{p-2}{p}}.
\end{eqnarray*}
\end{small}\noindent
Substituting it to the $\mu$ and $\mathbf{w}$, we obtain the solutions in Proposition \ref{prop:pmkl}.
The proof is complete.
\end{proof}

\bibliographystyle{plain}
\bibliography{mtl-tnn}

\end{document}